\documentclass{article} %
\usepackage{microtype}
\usepackage{hyperref}
\usepackage{url}
\usepackage{booktabs}
\usepackage{xspace}
\usepackage{changebar}
\usepackage{subcaption}  %

\usepackage[margin=1.0in]{geometry}
\usepackage[font=small,labelfont=bf]{caption} 
\usepackage[numbers,sort&compress]{natbib}

\usepackage{lineno}

\usepackage{amsmath,amssymb} %
\usepackage{enumitem}
\usepackage{cleveref}
\usepackage{graphicx}
\usepackage{multirow}
\usepackage{tcolorbox}
\tcbuselibrary{listings}
\tcbuselibrary{breakable}

\definecolor{darkblue}{rgb}{0, 0, 0.5}
\hypersetup{colorlinks=true, citecolor=darkblue, linkcolor=darkblue, urlcolor=darkblue}
\newcommand{\ours}
{\textsc{UniRG-CXR}\xspace}
\newcommand{\oursgeneral}
{\textsc{UniRG}\xspace}

\title{Scaling medical imaging report generation with \\multimodal reinforcement learning}

\author{Qianchu Liu\thanks{Equal contributions} ,
Sheng Zhang\footnotemark[1] ,
Guanghui Qin\footnotemark[1] ,\vspace{6pt}\\
Yu Gu, Ying Jin, Sam Preston, Yanbo Xu, Sid Kiblawi,\\
Wen-wai Yim, Tim Ossowski, Tristan Naumann, Mu Wei\thanks{Corresponding authors: muhsin.wei@microsoft.com, hoifung@microsoft.com} , ~Hoifung Poon\footnotemark[2]\\
Microsoft Research
}

\date{}

\begin{document}

\maketitle

\begin{abstract}

Frontier models have demonstrated remarkable capabilities in understanding and reasoning with natural-language text, but they still exhibit major competency gaps in multimodal understanding and reasoning especially in high-value verticals such as biomedicine. 
Medical imaging report generation is a prominent example. 
Supervised fine-tuning can substantially improve performance, but they are prone to overfitting to superficial boilerplate patterns.
In this paper, we introduce Universal Report Generation (\oursgeneral) as a general framework for medical imaging report generation.
By leveraging reinforcement learning as a unifying mechanism to directly optimize for evaluation metrics designed for end applications, \oursgeneral can significantly improve upon supervised fine-tuning and attain durable generalization across diverse institutions and clinical practices.
We trained \ours on publicly available chest X-ray (CXR) data and conducted a thorough evaluation in CXR report generation with rigorous evaluation scenarios. 
On the authoritative ReXrank benchmark,  \ours sets new overall SOTA, outperforming prior state of the art by a wide margin. We release our model at \url{https://huggingface.co/microsoft/UniRG-CXR}.

\end{abstract}

\section*{Main}

Medical imaging report generation has been an important application area for medical foundation models, aiming to automatically produce coherent and clinically meaningful diagnostic reports from medical images such as chest radiographs. Beyond its potential to reduce reporting burden, improve workflow efficiency, and enhance diagnostic consistency, report generation represents a key benchmark for evaluating broader multimodal reasoning capabilities in healthcare AI. Despite recent progress driven by large-scale vision–language models and domain-specific training corpora \citep{medgemma, medversa, llava-rad}, enabling models to produce faithful, clinically grounded reports that generalize across real-world imaging environments remains a substantial challenge.

A central obstacle lies in cross-institution generalization. For example, radiology reporting practices vary widely across datasets and healthcare systems, influenced by differences in institutional guidelines, departmental conventions, radiologist writing styles, and patient populations \citep{delbrouck2025automated}. Consequently, models trained through supervised fine-tuning (SFT) tend to inherit the lexical biases, phrasing patterns of their training datasets. While such models may achieve high scores on in-distribution benchmarks, they often show substantial degradation when evaluated on unseen institutions or external datasets. This brittleness is particularly concerning for safety-critical applications, where reliable performance across demographic groups, institutions, and imaging conditions is essential.

A second major limitation of prior work is that SFT objectives primarily optimize next-word prediction, which encourages surface-level lexical similarity to reference reports rather than alignment with clinically important factual attributes. As a result, conventional report generation systems often overfit to n-gram–based metrics (e.g., BLEU, ROUGE) that are only weakly correlated with radiological correctness \citep{yu2022external, llava-rad}. This misalignment highlights the need for training paradigms that optimize models directly for clinical usefulness and factual accuracy rather than superficial linguistic similarity.

To address these challenges, we introduce \oursgeneral, a unified and universal report generation framework built upon a novel reinforcement learning framework designed to enhance generalization, factual alignment, and robustness. Rather than relying on a generic SFT+RL pipeline, our method is designed to optimize an arbitrary weighted collection of clinically grounded objectives through a two-step optimization strategy, a composite reward aligned with RadCliQ, and a KL-regularized second stage for error reduction.

Using this framework, we train \ours on chest X-rays and show that it moves beyond dataset-specific reporting conventions, instead learning generalizable representations that yield consistently strong performance across diverse clinical settings.

We conduct extensive and rigorous evaluations to benchmark \ours against existing radiology report generation models across a broad spectrum of settings. Our evaluation suite includes standard report-level generation metrics, condition/diesease-level classification, cross-dataset/institution/demographics generalization, longitudinal evaluation and human evaluation reflecting real-world clinical practices. Across all axes, \ours achieves overall state-of-the-art (SOTA) performance, consistently surpassing prior baselines and demonstrating universal, reliable capabilities rarely observed in prior RRG systems.

In summary, our contributions are threefold:

{\bf A multi-objective reinforcement learning framework (\oursgeneral) for radiology report generation}, yielding \ours, a single model that attains overall SOTA performance across multiple datasets and evaluation metrics.

{\bf A comprehensive, clinically grounded evaluation of UniRG-CXR}, spanning report-level metrics, disease-level correctness, cross-dataset and cross-institution generalization, robustness analyses, longitudinal assessment and human evaluation.

{\bf Demonstration of \ours's universal capabilities}, showing that \ours overcomes the longstanding specialization and metric-overfitting issues of SFT-based report generation models by producing clinically aligned, generalizable reports across diverse real-world conditions.

Together, these advances establish UniRG as a robust and generalizable foundation for radiology report generation and highlight the promise of reinforcement learning as a key component of next-generation clinical vision–language models.

\section*{Results}

\begin{figure}[!ht]
    \centering
    \includegraphics[width=0.88\textwidth]    {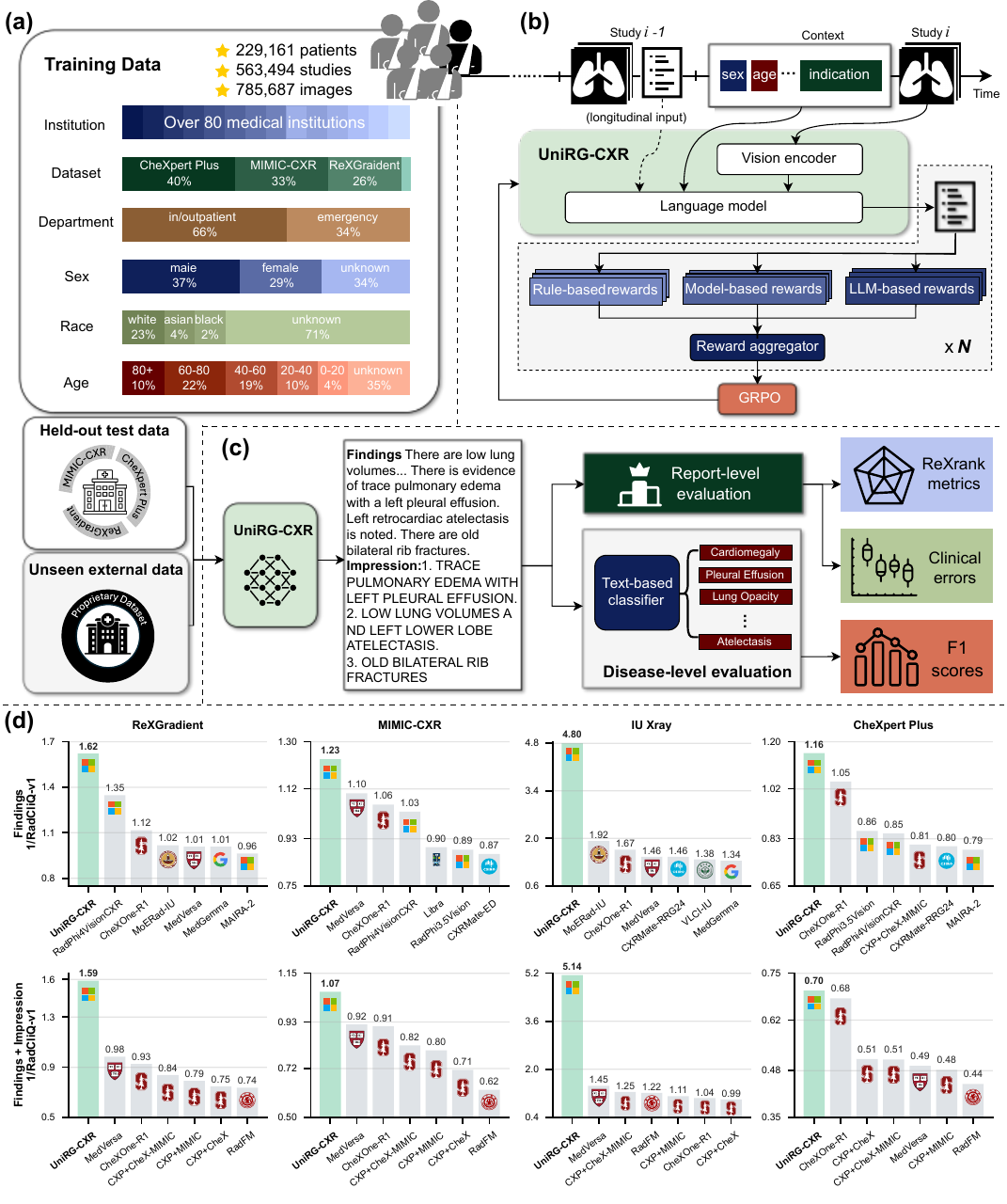}
    
    \caption{Overview of \ours. (a) Training Data: \ours is trained on the training splits of MIMIC-CXR~\citep{mimic-cxr}, CheXpert Plus~\citep{chexpert-plus}, ReXGradient-160k~\citep{rexgradient} and IU~\citep{iuxray} covering diverse institutions and patient demographics. 
(b) Training and Rewards: Taking input from the current image, clinical context (e.g., indication), and optionally prior studies, \ours uses GRPO reinforcement learning to optimize composite rewards that combine rule-based, model-based, and LLM-based metrics.
(c) Evaluation: We assess \ours on held-out test sets from MIMIC-CXR, CheXpert Plus, ReXGradient and additionally assess zero-shot generalization on an external proprietary dataset and IU-Xray (the zero-shot setting excludes IU-Xray training data). Report quality is measured using ReXrank metrics~\citep{rexrank} and an LLM-based clinical-error metric~\citep{zambrano2025clinically}, while diagnostic ability is evaluated via F1-based disease classification from generated reports.
(d) ReXrank Results: \ours achieves SOTA performance across four datasets and two generation settings (findings only and findings + impression), showing substantial gains over prior state-of-the-art.
\protect
All statistical significance tests are reported in \Cref{sec:stats-test}.
\protect
}
    \label{fig:overview-of-unirg-cxr}
\end{figure}

\subsection*{Overview of UniRG and \ours}
We propose \oursgeneral, a next-generation approach for medical imaging report using reinforcement learning. With this framework, we train \ours, a state-of-the-art model for radiology report generation that produces clinically faithful reports across diverse institutions and datasets with substantial lead over existing baselines. Built upon the open-source Qwen3-VL-8B-Instruct foundation, \ours combined supervised fine-tuning with reinforcement learning to directly optimize for clinically relevant objective aggregating multiple metrics spanning rule-based (BLEU), model-based (BERTScore, SembScore, RadGraph F1), and LLM-based (CheXprompt) metrics. Trained on large-scale datasets including MIMIC-CXR, CheXpert Plus, ReXGradient and IU (covering over 560k studies from 80+ institutions), \ours can condition on the current image, contextual text, and prior study information where available and output both findings and impression as a full radiology report from the x-ray interpretation. To evaluate the quality of \ours, we assess \ours on held-out test sets (MIMIC-CXR~\citep{mimic-cxr}, CheXpert Plus~\citep{chexpert-plus}, ReXGradient~\citep{rexgradient}, IU~\citep{iuxray}) and unseen proprietary data. Report quality is measured using ReXrank metrics~\citep{rexrank} and CheXprompt, an LLM-based clinical-error metric~\citep{zambrano2025clinically}, while diagnostic ability is evaluated via F1-based disease classification from generated reports. \ours demonstrates unprecedented generalization and robustness, outperforming prior state-of-the-art systems (e.g., MedVersa, MedGemma, MAIRA-2) by substantial margins across public and private benchmarks across our evaluation settings.

As shown in (d) and (e) from \Cref{fig:sota-report-gen}, we present the detailed ReXrank leaderboard results\footnote{Results are collected from the ReXrank leaderboard \citep{rexrank} on Feb 2026} which compares \ours against previous SOTA models across four datasets—ReXGradient \citep{rexgradient}, MIMIC-CXR \citep{mimic-cxr}, IU X-Ray \citep{iuxray}, and CheXpert Plus \citep{chexpert-plus}. Each dataset is evaluated under two generation settings: findings-only and the more challenging findings + impression setup. Performance is measured by 1/RadCliQ-v1 \citep{yu2023evaluating} (higher is better), a composite metric used as the default metric in RexRank leaderboard and was found to correlate more strongly with human judgment than individual metrics. As shown in \Cref{fig:overview-of-unirg-cxr}, \ours achieves consistent and substantial gains over prior SOTA models in every steup.  Notably, on the ReXGradient and IU test set, \ours exceeds the previous best model by over 50\%, underscoring its strong generalization capability for radiology report generation.
\subsection*{\ours achieves SOTA with universal improvements across metrics}

\begin{figure}[!ht]
    \centering
    \includegraphics[width=\textwidth]
    {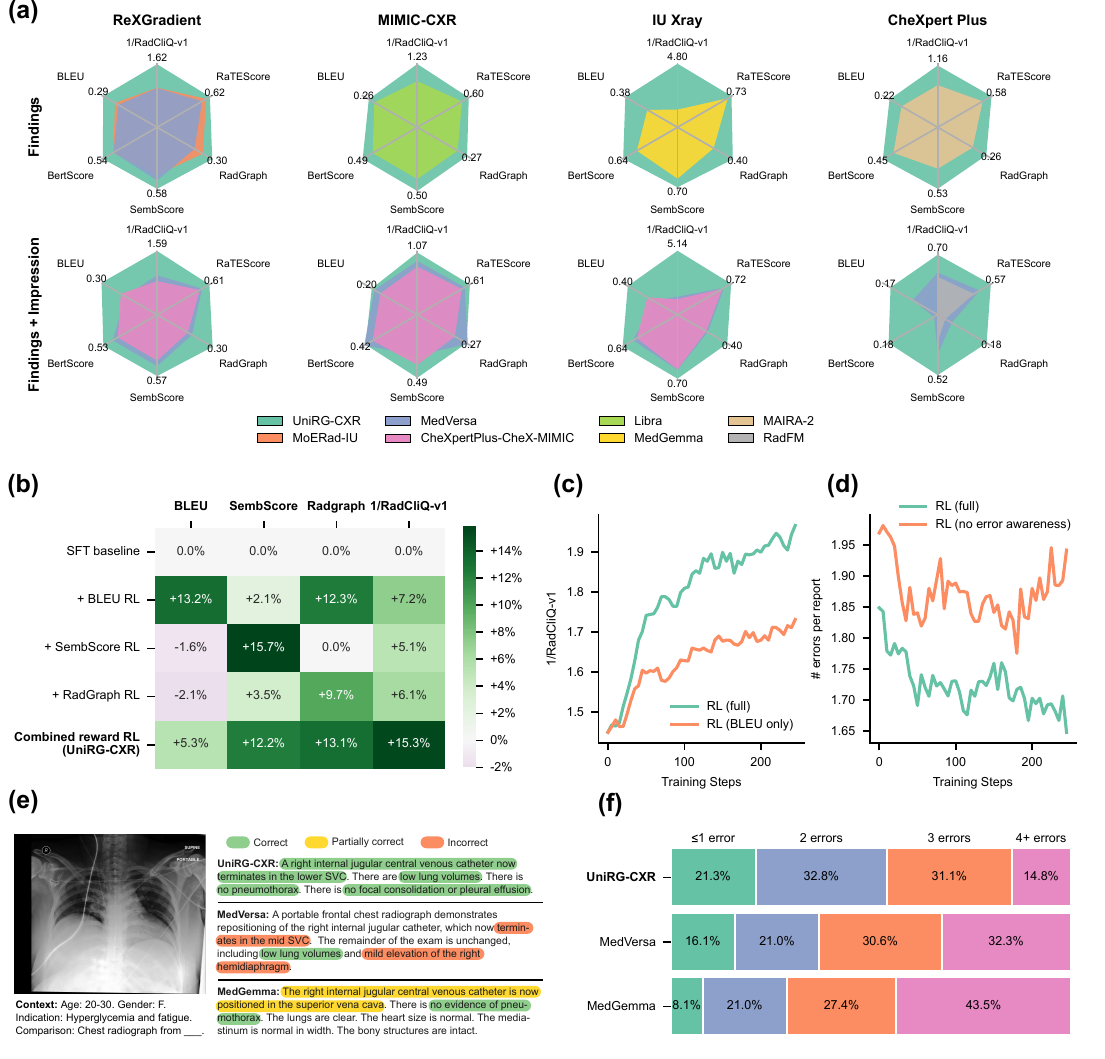}
    \caption{\ours achieves state-of-the-art performance, delivering consistent and comprehensive performance gains across metrics. (a) On the ReXrank leaderboard, \ours (green) shows robust, universal improvement across all evaluation metrics.  (b). Starting from the same SFT checkpoint, RL with our combined reward achieves more balanced gains across metrics and the highest RadCliQ-v1 score compared to RL on single metrics. This ablation study is trained and tested on MIMIC (c). Ablation study on the training dynamics shows RL full (\ours) achieves significantly better RadCliQ-v1 score than RL only on BLEU. (d). During training, RL full (\ours) shows a steady decrease in clinical errors per report as compared with a fluctuating trajectory without consistent improvement from an ablation run without error awareness (i.e. removing CheXprompt metric optimization). Both (c) and (d) show results on 1024 MIMIC validation set from ablations that are trained on MIMIC. (e). Case studies illustrate that \ours can produce error-free reports, unlike MedVersa and MedGemma. (f). \ours yields a substantially higher proportion of reports with $\leq 1$ error and fewer with $\geq 4$ errors
    than prior models.
    \protect
    All statistical significance tests are reported in \Cref{sec:stats-test}.
    \protect
 }
    \label{fig:sota-report-gen}
\end{figure}

While RadCliQ-v1 serves as our primary evaluation metric, \Cref{fig:sota-report-gen} further illustrates that \ours delivers broad improvements across diverse metrics including both those emphasizing lexical similarity (e.g., BLEU, BERTScore) and those emphasizing factual correctness (e.g., SembScore). Notably, \ours also improves metrics it was not explicitly optimized for, such as RaTEScore, highlighting that its performance gains are universal and multi-faceted rather than overfitted to specific metrics (\Cref{fig:sota-report-gen} (a)). The key to this universal improvements lies in our combined reward RL which jointly optimizes multiple individual metrics. As shown in \Cref{fig:sota-report-gen} (b) and (c), we compare \ours with ablation studies that only optimize an individual metric. We show that our proposed combined reward RL achieves balanced gains across individual metrics and yields overall the best RadCliQ-v1 score.

Beyond the ReXrank-provided metrics, we additionally evaluate \ours using CheXprompt \citep{zambrano2025clinically}, an LLM-based metric that quantifies the number of clinical errors relative to the reference report (following \citet{zambrano2025clinically}, we use GPT-4 as the backbone evaluator). As shown in \Cref{fig:sota-report-gen} (f), we compare the proportion of generated reports with varying numbers of clinical errors ($\leq 1$, 2, 3, and $\geq 4$) across \ours, MedVersa, and MedGemma. UniRG-CXR produces substantially more error-free or low-error reports (21.3\% $\leq 1$ error) compared with prior state-of-the-art systems (Medversa 16.1\% and MedGemma 3.1\%), while markedly reducing the fraction of high-error reports ($\geq 4$ errors: 14.8\%) relative to MedVersa (32.3\%) and MedGemma (43.5\%). These results indicate that \ours achieves more clinically faithful and accurate report generation. Qualitatively, we show an example where \ours can generate error-free report that covers all the important findings while output from MedGemma and MedVersa contain errors or partial errors. The key to the improvement on clinical error reduction is the incorporation of error awareness metric (i.e. CheXprompt) in our combined RL recipe in \ours. In \Cref{fig:sota-report-gen} (d), the RL(no error awareness) ablation that does not contain CheXprompt optimization shows stagnant learning in terms of reducing clinical errors whereas our full RL recipe from \ours that incorporates error awareness leads to a steady downward trajectory of reduction in report errors throughout training, indicating that explicit optimization for clinical correctness effectively improves report fidelity.

\subsection*{\ours enhances longitudinal report generation}

\begin{figure}[!ht]
    \centering
    \includegraphics[width=\textwidth]
    {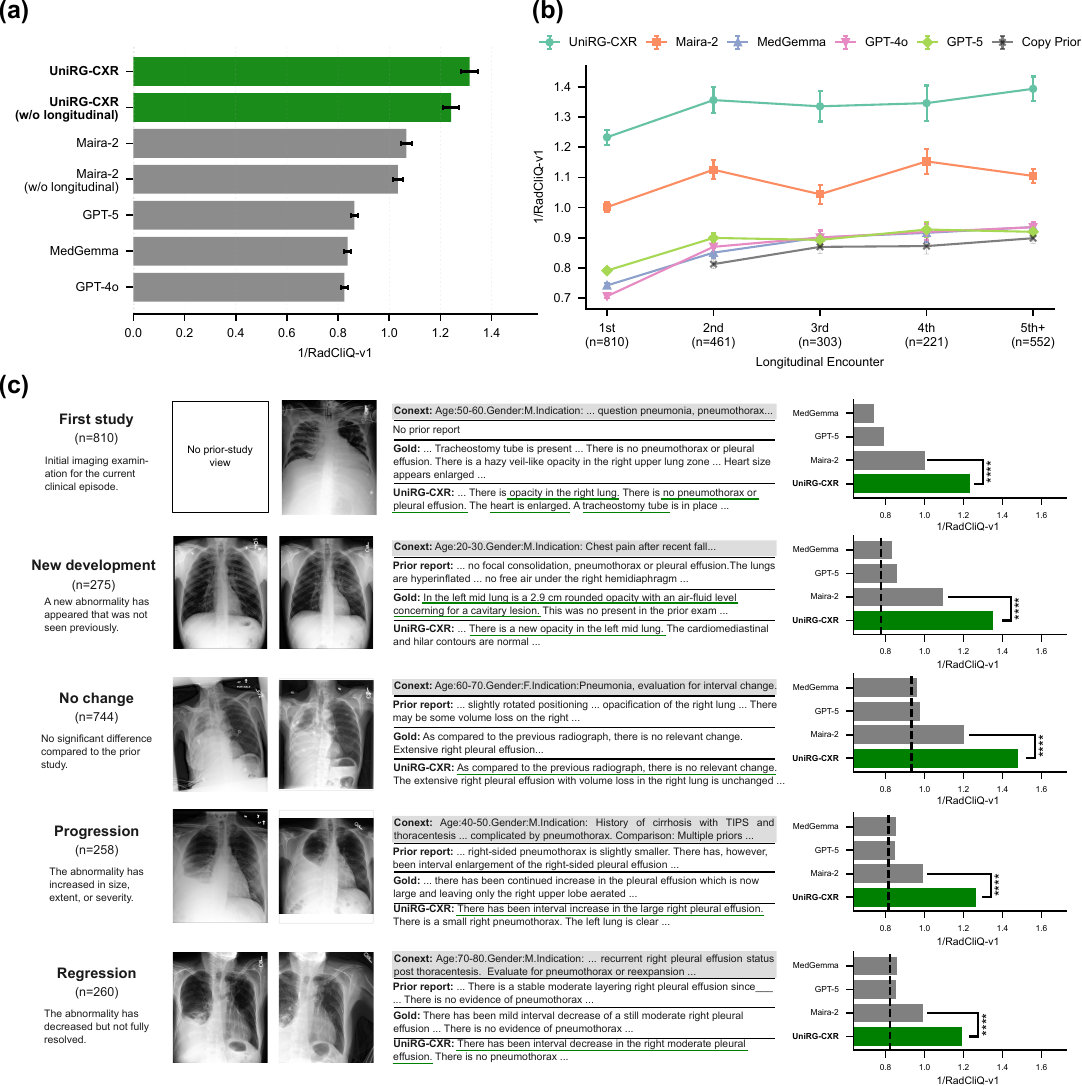}
    \caption{\ours enhances longitudinal report generation. (a). Comparing \ours and its non-longitudinal ablation with prior models on longitudinal report generation, we show \ours exhibits the best performance and the longitudinal information is beneficial to the performance. (b). \ours achieves the best performance across different longitudinal encounter points ranging from the first encounter to the more complex 5th+ encounters, showcasing its improvements are across the board. In comparison, prior models such as GPT-5, GPT-4o and MedGemma are barely surpassing the copy prior report baseline (grey lines).  (c). Compared with prior models which barely improve over the copy prior baseline (dashed line), \ours significantly and consistently improves performance across different temporal disease change categories including new development, no change, progression and regression (categorized by GPT-5 on ground truth report). Qualitative examples are shown for each category where \ours correctly predicts the temporal change based on the input. All results in this figure are on MIMIC test set with prior information where available.
    \protect
    All statistical significance tests are reported in \Cref{sec:stats-test}.
    \protect
    }
    \label{fig:longitudinal}
\end{figure}

In the real-world setting, radiologists routinely reference prior studies (both reports and images) when interpreting the current exam, often noting changes such as whether pneumonia has improved or worsened compared to a previous scan. To better approximate this realistic workflow, we incorporate longitudinal training, enabling the model to condition on both the prior image and prior report, during the RL training. As shown in \Cref{fig:longitudinal}, \ours achieves the best performance in longitudinal report generation compared with prior longitudinal report generation models such as Maira-2 and frontier large language model such as GPT-5. We also notice that longitudinal information is effectively incorporated in \ours as it boosts performance over its non-longitudinal set up. To understand the gain of \ours in more granularity, we further categorize the reports as shown in \Cref{fig:longitudinal} (b). The test studies from MIMIC are split into five categories according to their encounter time points ranging from first encounter report without prior information to increased complexity in 2nd, 3rd, 4th and 5th+ encounter points where the report is written in reference with multiple encounters the patient has experienced in the history. We observe that the first-encounter reports are generally the most challenging, and as the number of encounters increases, report quality improves consistently. This trend is intuitive: for the first encounter, the model must generate a completely new description based solely on the current image, without any prior context. In contrast, subsequent encounters provide previous reports that capture the patient’s underlying conditions, enabling the model to generate more accurate and contextually faithful reports. Across all encounter points, \ours achieves substantial performance gains over prior models. Moreover, \ours significantly outperforms the “copy prior report” baseline, demonstrating that it effectively leverages prior information rather than relying on it as a shortcut, a behavior observed in some competing models (e.g., MedGemma, GPT-4o, and GPT-5) which only marginally exceed the copy-prior baseline. As longitudinal report generation involves capturing temporal changes in disease state across patient encounters. In \Cref{fig:longitudinal}(c), we categorize each report into five temporal description types (first study (no prior study), new development, no change, progression, and regression) using GPT-5 as an automatic labeler. We evaluate \ours’s performance across these categories, comparing it against prior models and the copy prior report baseline. As expected, the no change category is the easiest, since much of the content can be reused from the prior report. In contrast, categories reflecting disease evolution such as regression or new development are more challenging, as the model must accurately localize and quantify subtle changes. \ours demonstrates consistent and substantial improvements across all categories, markedly surpassing both prior models and the copy-prior baseline. Qualitative examples further illustrate that \ours generates clinically faithful longitudinal descriptions, correctly identifying changes such as new findings or resolving abnormalities. Collectively, these results highlight \ours’s superior capability to model and reason over longitudinal patient trajectories.

\subsection*{Generalization and Robustness of \ours}

\begin{figure}[!ht]
    \centering
    \includegraphics[width=\textwidth]
    {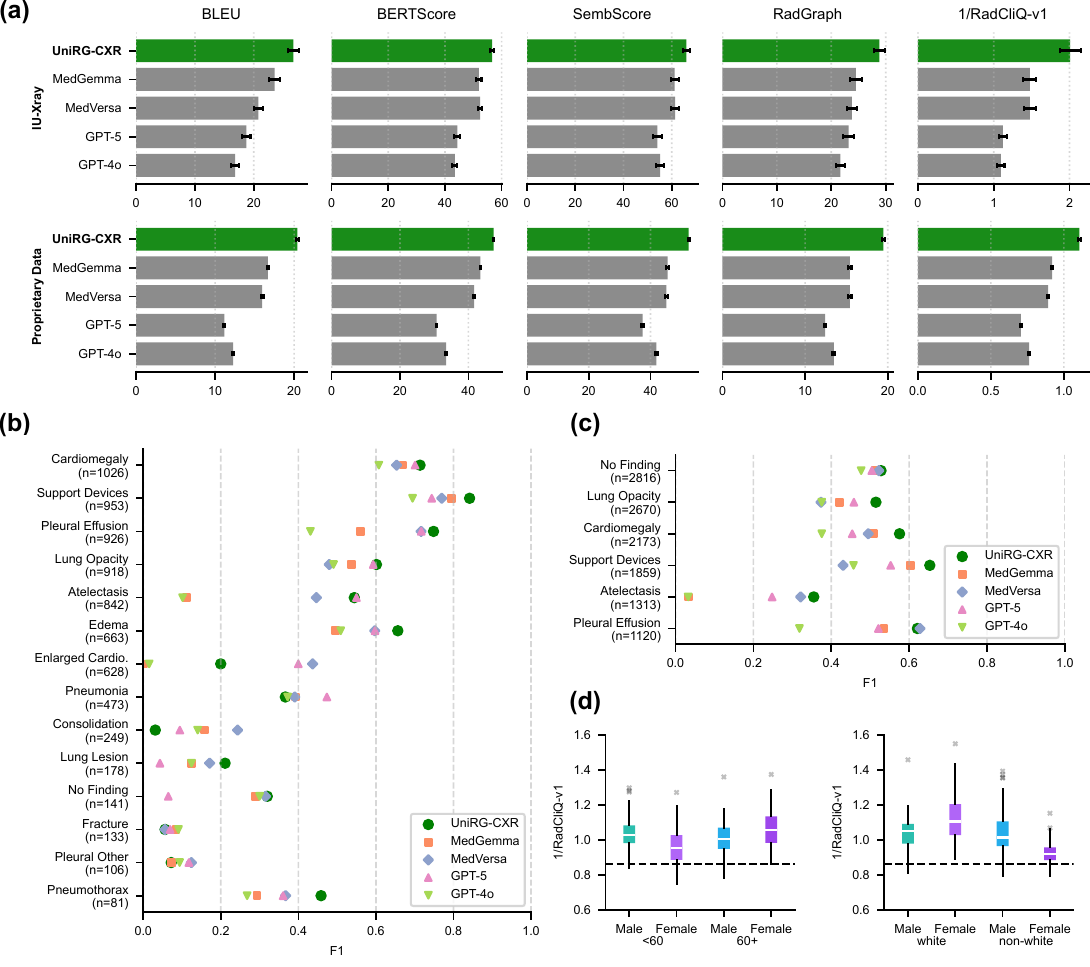}
    \caption{Generalization and robustness of \ours. (a). We held out two datasets sources (IU-Xray and PD (proprietary data) from the training data and evaluate \ours in a zero-shot setting on these datasets . \ours consistently outperforms prior models, maintaining substantial performance gains in this challenging setup. (b) and (c) present condition-level F1 scores on MIMIC-CXR and PD and highlight that \ours remains the overall top-performing model in condition-level diagnostic accuracy. (d). \ours demonstrates stable and robust performance across gender, age, and race subgroups, all of which exceed the performance of the second-best model (the dashed lines).  %
    \protect
    All statistical significance tests are reported in \Cref{sec:stats-test}.
    \protect
    }
    \label{fig:generalizability}
\end{figure}

In this section, we evaluate the robustness and generalization of \ours. First of all, a robust universal report generation model should generalize well across institutions, including those with data distributions unseen during training. To test this, we create an experiment setup where we intentionally leave out certain data sources (IU data and a propriety data source) from the training of \ours. We then test the model's zero-shot performance on these two datasets, as shown in \Cref{fig:generalizability} (a). \ours consistently achieves the best performance across all out-of-distribution datasets, surpassing prior baselines with substantial margins.

Another key generalization of chest radiology report generation models is to accurately identify and classify specific thoracic diseases/conditions, thereby expediting the diagnostic process for clinicians~\citep{milam2023current}. We evaluate the condition diagnosis capabilities of the output from \ours by applying the CheXbert model \citep{chexbert} to detect diseases from its generated reports. As shown in \Cref{fig:generalizability} (b) and (c), \ours is leading the performance across all the prevalent diseases compared with other prior models in both MIMIC and out-of-distribution proprietary dataset. 

Finally, we evaluate the robustness of \ours across gender, age, and race. As shown in \Cref{fig:generalizability}(d), we present performance stratified by these demographic subgroups on the CheXpert-Plus dataset. \ours achieves consistently high scores across all groups, with overlapping distributions and no noticeable performance drop in any demographic category. Moreover, all subgroup performances of \ours surpass those of the second-best model (dashed line), demonstrating that \ours is robust and fair across demographic subgroups while maintaining superiority over prior models.

\protect
\subsection*{Human Evaluation in a Radiologist-Centric Workflow with Failure Mode Analysis}
To further validate the performance from \ours, we conducted an external real-world clinical validation study by radiologists (\Cref{tab:combined_eval} (A)). Specifically, our annotation test set comes from the CheXpert-Plus dataset following \citep{yu2023evaluating}. Four radiologists were recruited to assess each report from UniRG-CXR, GPT-5, MedGemma, Medversa. The input for the annotators is the current and previous xray scans together with any available context such as indication. We asked the annotators to rate \textit{completeness}, \textit{factual accuracy} and \textit{overall} on a scale of 1 - 5  as well as provide \textit{error annotation}. Overall inter-annotator agreement is 0.54 (\Cref{appendix: iaa} shows detailed breakdown of inter-annotator agreement across metrics).
The results verified the superiority of \ours as \textbf{UniRG-CXR is the most preferred model by human radiologists}, achieving the highest ratings while showing the lowest number of errors ((\Cref{tab:combined_eval} (B)). These expert validation results further corroborated the core findings in our paper. 

To better understand the performance of \ours and the baselines, we asked annotators to categorize errors into clinically meaningful failure modes. We focus on two categories: (1) clinically significant false predictions, where the model generates findings that contradict the X-ray image, often due to hallucination or misinterpretation; and (2) clinically significant omissions, where the model fails to report important findings that are evident in the image. \Cref{tab:combined_eval}(C) presents the detailed breakdown for these error categories. We observe that \ours consistently yields substantially fewer false predictions and omissions compared to the baselines. Nevertheless, \ours is not error-free, underscoring the importance of radiologist oversight and validation in real-world deployment. \Cref{appendix: error} shows qualitative examples for each error category from \ours and the other baselines.

\begin{figure}[!ht]
\centering
\small
\includegraphics[width=\textwidth]{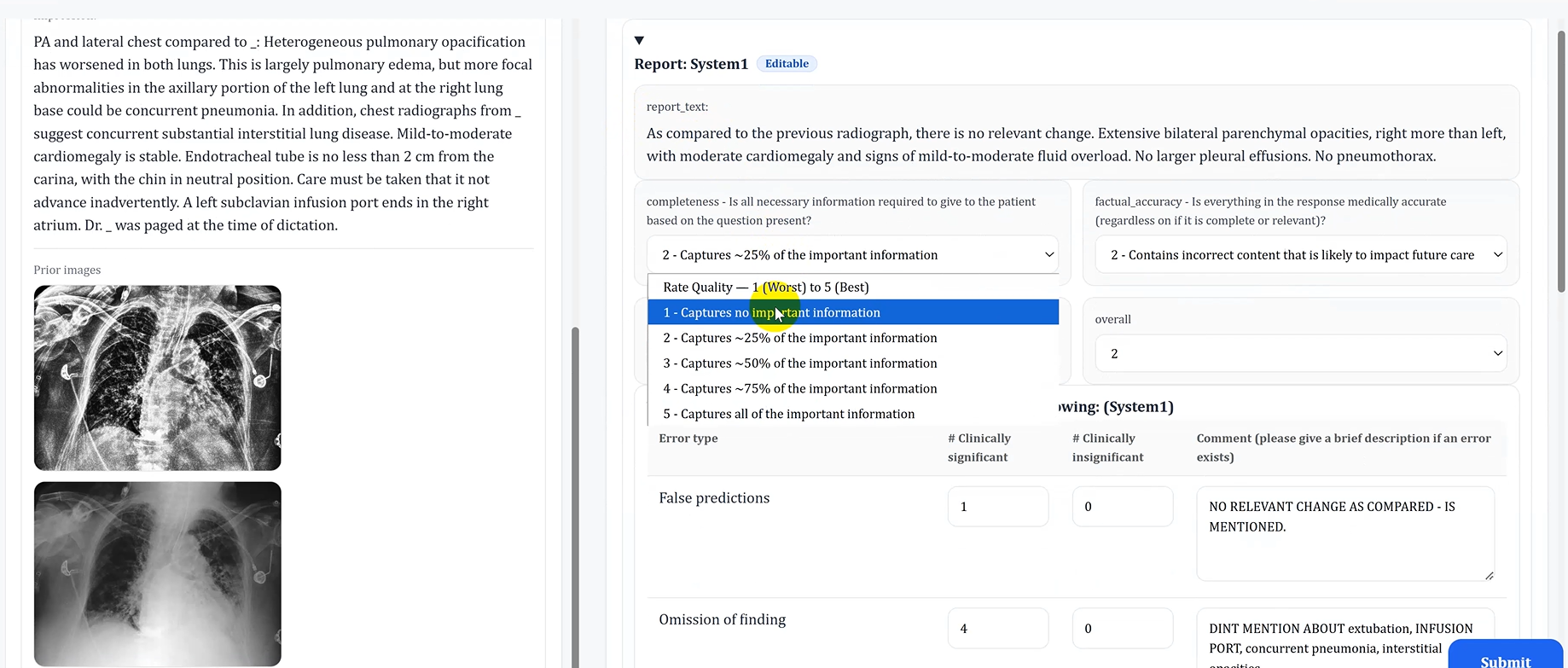}
\caption{Annotator's UI interface for evaluating each system's radiology report generation output \label{fig:annotator_ui_for_r1}}
\end{figure}

We also believe that successful real-world deployment requires careful integration with existing radiology workflows, along with thoughtful user interface design, explainability, and radiologist acceptance. During our expert radiologist evaluation, we iteratively refined the annotation interface and instructions based on pilot feedback to improve usability and ensure reliable report assessment. A screenshot of the annotation interface is shown in \Cref{fig:annotator_ui_for_r1}. The interface is designed to mirror clinical workflow, presenting the model-generated output as a draft report alongside all relevant context. Radiologists are asked to evaluate key quality metrics and provide free-text comments for additional concerns. This interaction can be viewed as a human-in-the-loop verification step for reviewing machine-generated reports.

\begin{table*}[t]
\centering
\small
\caption{Expert radiologist evaluation setup and results. Lower is better for False pred., Omission, and Total Errors; higher is better for other metrics.}
\label{tab:combined_eval}
\setlength{\tabcolsep}{8pt}
\renewcommand{\arraystretch}{1.1}

\begin{tabular}{lcccc}
\toprule

\addlinespace[3pt]
\multicolumn{5}{c}{\textbf{(A) Annotator Information}} \\
\addlinespace[3pt]
\midrule
\multicolumn{2}{l}{\textbf{\# Annotators}} & \multicolumn{3}{l}{4} \\
\multicolumn{2}{l}{\textbf{Countries}} & \multicolumn{3}{l}{2 US, 1 Canada, 1 India} \\
\multicolumn{2}{l}{\textbf{Requirements}} & \multicolumn{3}{p{9cm}}{Board-certified in their country of practice with 5--10 years of experience.} \\
\multicolumn{2}{l}{\textbf{Time spent}} & \multicolumn{3}{l}{3 months} \\
\multicolumn{2}{l}{\textbf{Platform}} & \multicolumn{3}{l}{Centific} \\

\addlinespace[8pt]
\midrule
\midrule
\addlinespace[3pt]
\multicolumn{5}{c}{\textbf{(B) Human Evaluation Results}} \\
\addlinespace[3pt]
\midrule
\textbf{Model} & \textbf{Overall} $\uparrow$ & \textbf{Completeness} $\uparrow$ & \textbf{Factual Accuracy} $\uparrow$ & \textbf{Total Errors} $\downarrow$ \\
\midrule
MedGemma  & 3.41 & 3.38 & 3.81 & 3.33 \\
MedVersa  & 3.56 & 3.54 & 3.88 & 3.11 \\
GPT-5     & 3.77 & 3.81 & 3.77 & 2.58 \\
\midrule
\ours     & \textbf{3.92} & \textbf{3.98} & \textbf{4.32} & \textbf{2.49} \\

\addlinespace[8pt]
\midrule
\midrule

\addlinespace[3pt]
\multicolumn{5}{c}{\textbf{(C) Further categorizing clinically significant errors}} \\
\addlinespace[3pt]
\midrule
\multicolumn{5}{c}{
\begin{tabular*}{0.82\textwidth}{@{\extracolsep{\fill}}lcc}
\textbf{Model} & \textbf{False pred.} $\downarrow$ & \textbf{Omission} $\downarrow$ \\
\midrule
GPT-5     & 0.492 & 0.975 \\
MedGemma  & 0.533 & 1.283 \\
MedVersa  & 0.433 & 1.183 \\
\midrule
\ours     & \textbf{0.292} & \textbf{0.875} \\
\end{tabular*}
} \\

\bottomrule
\end{tabular}
\end{table*}

\protect

\section*{Discussion}
In this work, we introduce \oursgeneral, a reinforcement learning-based framework for medical imaging report generation. Using this framework, we are able to train \ours, a state-of-the-art radiology report generation system for chest radiographs,
which establishes a new performance standard across benchmarks on the ReXrank leaderboard, 
which spans multiple clinical contexts (inpatient, outpatient, and emergency care) and data from over 70 medical sites.
Unlike previous models, which often excelled on isolated benchmarks or on specific metrics, UniRG-CXR delivers consistent performance improvement across multiple metrics on four widely used datasets (MIMIC-CXR, IU-Xray, CheXpert Plus, and ReXGradient) as well as additional proprietary dataset.
The optimization framework of \oursgeneral{} supports an arbitrary weighted combination of verifiable, clinically grounded objectives tailored to report-generation use cases.
In this study, we instantiate and evaluate this framework in chest X-ray reporting, which provides large-scale public data and established benchmarks such as ReXrank~\citep{rexrank}.

Our evaluation protocol provides a comprehensive assessment of AI-based radiology report generation. We benchmark the system using lexical similarity, embedding-based similarity, and LLM-based clinical error metrics, enabling a holistic comparison against both reference reports and competing models. Beyond these, we conduct longitudinal evaluations that simulate real-world radiologists’ workflows, assess generalization and robustness on unseen distributions, and stratify performance by demographic subgroups. We further link report-level quality to condition-level diagnostic accuracy.
Finally, we conduct human evaluation to verify the superiority of \ours over baselines.
Collectively, these analyses underscore the universality, robustness, and clinical alignment of \ours.

\paragraph{Universal across institutions and data distributions}
Radiology reporting practices vary widely across institutions, regions~\citep{hartung2020create}, and documentation conventions~\citep{kahn2009toward}.
Prior systems often overfit to dataset-specific phrasing or reporting styles, leading to performance drops when evaluated on unseen datasets or across sites~\citep{huang2023generative,nicolson_improving_2023,tanida2023interactive}.
\ours overcomes this limitation by exhibiting consistent high performance across all benchmark datasets and on entirely unseen distributions with zero-shot inference, including both public and private cohorts collected from diverse institutions.
These results indicate that \ours captures the underlying clinical semantics rather than memorizing superficial textual templates—achieving true generalization across data sources, institutions, and domains.
Such robustness establishes \ours as a universal foundation model for radiology reporting, capable of maintaining reliability and fidelity across heterogeneous real-world environments.

\paragraph{Universal across evaluation metrics}

Traditional text-generation metrics such as BLEU or ROUGE correlate only weakly with clinical judgments~\citep{liu2019clinically,radcliq}, often obscuring medically significant errors. \ours bridges this gap by explicitly integrating clinical error signals into its reinforcement learning reward design, aligning optimization with radiological practice rather than surface-level linguistic similarity. As a result, it achieves strong and balanced performance across both NLG and clinically grounded metrics, representing a universal improvement across evaluation dimensions.

\paragraph{Universal across diagnostic levels}
Prior work typically assessed performance at the report level~\citep{medversa,medgemma,llava-rad}, obscuring whether models captured critical findings~\citep{milam2023current}.
By incorporating disease-level assessments, we provide a finer-grained view of diagnostic fidelity.
These evaluations showcase the ability of \ours to reflect diagnostic information, including in the long tail of rare conditions where training data are sparse.

\paragraph{Universal across longitudinal contexts}
Radiologists often rely on prior studies when interpreting new exams~\citep{bannur2023learning,serra2023controllable,zhu2023utilizing,maira-2}. 
While previous models primarily operated in single-study settings~\citep{medversa}, we additionally evaluate a longitudinal setting in which the model must integrate not only current CXRs but also prior reports and images.
\ours excels in both standard and longitudinal conditions. It effectively integrates prior images and reports to produce clinically coherent updates, achieving state-of-the-art results in temporal reasoning and demonstrating superior longitudinal modeling capability.
Here, multimodal reasoning refers to the integration of heterogeneous clinical inputs for clinically faithful report generation, rather than explicit chain-of-thought generation or region-level visual grounding.

\paragraph{Universal across demographics}
We further validate \ours across stratified demographic subgroups, confirming robust and equitable performance across gender, age, and race. This fairness evaluation is critical for real-world deployment, ensuring minimal bias and consistent report quality across patient populations.

Overall, \ours represents a substantial advance in radiology report generation, unifying high performance across datasets, metrics, diagnostic levels, longitudinal setups, and demographic subgroups. Beyond surpassing prior systems, it embodies the principles of universality, generalizability, and clinical alignment, paving the way for reliable real-world deployment.
In the longer term, augmenting this system with interactive, instruction-following capabilities and expanding to multimodal patient data (e.g., lab tests, prior imaging, and clinical notes) could further enhance its clinical utility. 
At the same time, we acknowledge the present study is limited to chest X-ray report generation.
Extending the framework to additional imaging modalities and to richer evaluations of multimodal reasoning remains an important direction for future work.
We anticipate that \ours will serve as both a strong benchmark for future research and a foundation for building reliable, assistive AI systems in radiology.

\section*{Methods}

\subsection*{Model}
\ours is built by fine-tuning a state-of-the-art open-source vision–language foundation model, Qwen3-VL-8B-Instruct \citep{bai2025qwen3vltechnicalreport} for the report generation tasks. In the sections that follow, we describe our task formulation, inference and evaluation protocols, optimization strategy, and dataset curation in detail.
\subsection*{Tasks}
Our model is trained to generate both the ‘findings’ and ‘impression’ sections of the report for a frontal view (anterior–posterior or posterior–anterior) of the chest radiograph, which typically capture the key observations made in a study. The model receives additional contextual information, including the study indication and any available comparison text. To improve computational efficiency, each radiograph is resized to a resolution of 512 × 512 pixels. In routine clinical practice, radiologists frequently reference prior images and prior reports when interpreting the current study. To mirror this workflow and enhance clinical fidelity, we also supply the model with the most recent prior frontal radiograph and its associated report as supplementary inputs.

\subsection*{Inference and Evaluation Metrics}
We follow the exact setup from ReXrank to take into account context (indication + comparison). We only use the key image path provided by ReXrank which is typically frontal view image. We keep temperature as 0 for inference. We follow the setups in ReXrank \citep{rexrank} to evaluate report quality using the following metrics: 

BLEU-2 \citep{papineni2002bleu}. BLEU is a standard metric for machine translation and text generation that measures n-gram precision between generated and reference texts (0–1 scale). Following \citet{rexrank}, we report BLEU-2, which captures bigram precision.

BERTScore \citep{zhangbertscore, devlin2019bert}. BERTScore evaluates semantic similarity by computing cosine similarity between BERT embeddings of the generated and reference reports, providing a meaning-aware alternative to surface n-gram metrics.

SembScore \citep{chexbert}. SembScore is a radiology-specific metric that computes cosine similarity between 14-pathology indicator vectors produced by the CheXbert labeler for generated and groundtruth reports.

RadGraph-F1 \citep{jain1radgraph} measures the overlap in clinical entities and relations extracted by RadGraph from candidate and reference reports.

1/RadCliQ-v1 \citep{yu2023evaluating} is the reciprocal of the RadCliQ composite metric, which aggregates BLEU, BERTScore, SembScore, and RadGraph-F1 for holistic radiology report evaluation. Because RadCliQ is originally lower-is-better, we follow ReXrank \citep{rexrank} and report its inverse so that higher values indicate better performance to be consistent with other scores. 

RaTEScore \citep{zhao2024ratescore} is an entity-centric metric emphasizing key medical concepts such as diagnoses and anatomical structures, while being robust to medical synonyms and negation. 

CheXprompt \citep{llava-rad} In addition to the ReXrank leaderboard metrics, we also evaluate report quality with CheXprompt, an LLM-based error detection metric. 
 
\subsection*{Optimization}

\ours uses a two-stage SFT + RL pipeline in which the reinforcement learning stage is designed to optimize a clinically grounded combination of objectives, rather than serving as a standard post-SFT refinement step.
The key contribution is not merely applying RL after SFT, but designing a task-specific multi-objective RL recipe that can balance clinically grounded rewards for report similarity, structured factual consistency, and clinical error reduction.

In the {\bf SFT stage}, the model is initialized with a strong foundation for radiology report generation.  The SFT training is performed over four datasets: MIMIC, CheXpert-Plus, ReXGradient and IU and we conduct a grid search over learning rates \([1\times10^{-5},\, 5\times10^{-5}]\) and batch sizes \([128,\,256,\,521]\). The optimal configuration is a learning rate of \(5\times10^{-5}\) with a batch size of 256 for 3 epochs.

In the {\bf RL stage}, we adopt GRPO~\citep{shao2024deepseekmath} as our reinforcement learning algorithm, which eliminates the need for value functions by computing advantages within query-specific groups. Following recent advances~\citep{yu2025dapo}, we incorporate two key improvements:  
(1) a higher clipping threshold to encourage response diversity and prevent entropy collapse;  
(2) remove KL penalty.
Our training adopts a learning rate of \(5\times10^{-6}\), a global batch size of 256, and 16 sampled rollouts per query. The RL training is performed over four datasets: MIMIC, CheXpert-Plus, ReXGradient and IU. 
For reward optimization in the RL stage, we target BLEU, BERTScore, RadGraph-F1, SembScore, and CheXprompt (LLM-based), without using any format rewards since the model reliably produces well-structured reports after SFT. Our RL procedure follows a two-step optimization strategy.

\textbf{Step 1: RadCliQ-oriented optimization.}
We first optimize a weighted composite reward consisting of BERTScore, SembScore, and RadGraph-F1, using the RadCliQ coefficients of 0.370, 0.253, and 0.377, respectively, following the formulation of RadCliQ~\citep{yu2023evaluating}.
All rewards are computed from the generated report and its corresponding ground-truth report. 
Because RadGraph-F1 is originally defined for batched evaluation, we adapt it to operate at the single-instance level for RL training.
This stage runs for one epoch and encourages the model to generate outputs that are both lexically and clinically aligned with the ground-truth reports, effectively targeting RadCliQ improvement.

\textbf{Step 2: Error-reduction optimization.}
Starting from the best checkpoint from Step~1's RadCliQ optimization, we perform an additional epoch in which we incorporate the CheXprompt error metric into the reward. Specifically, we use \(1 / (\#\text{ CheXprompt errors} + 1)\) as the CheXprompt reward to incentivize reducing factual reporting errors. To preserve the RadCliQ performance achieved in Step~1, we integrate the CheXprompt reward with a weight of 0.5 alongside the previous metrics. We also apply a KL regularization term with coefficient 0.03 to prevent excessive deviation from the Step~1 policy.
The KL constraint is critical in this stage because it preserves the gains from the first-stage policy while adding explicit error-awareness instead of over-optimizing the LLM-based reward.
Additional ablations in the Supplementary Information show that both a Dr.~Tulu-style LLM-as-a-judge RL setup~\citep{shao2025dr} and simpler variants without the full two-step composite-reward design underperform the final \ours recipe (\Cref{tab:llm_judge_comparison,tab:unirg_ablation_recipe}).
These results support the importance of the staged optimization design: optimizing a single reward or combining all objectives in a simpler formulation yields inferior overall performance.

\subsection*{Dataset Details}
Our training data consists of the training splits from MIMIC-CXR, CheXpert Plus, ReXGradient and IU. We extracted the indication, comparison, findings, and impression sections from the corresponding radiology reports. We then removed studies in which both the findings and impression sections were empty. Studies that contained a findings section but lacked an impression, or contained an impression but lacked findings, were retained. Depending on which ground-truth sections were available (findings and/or impression), we applied different prompt templates, as shown in the supplementary information section. We also set aside 1,024 samples from the MIMIC-CXR training set as a validation set for all experiments.

 We follow ReXrank evaluation to evaluate \ours on ReXrank official test sets from MIMIC-CXR, CheXpert Plus, IU-Xray and ReXGradient private test set. Apart from the ReXrank testsets, we also evaluate on a  proprietary dataset which we name as PD. Below are the details for each dataset. 
\paragraph{MIMIC-CXR} \citep{mimic-cxr}. A large, publicly available dataset containing 377,110 chest X-rays from 227,835 studies collected at the Beth Israel Deaconess Medical Center in Boston, MA. All images and reports are fully de-identified. We use the official ReXrank MIMIC test set, which includes 2,347 studies, for evaluation, and use the remaining training split for model development.

 \paragraph{CheXpert Plus} \citep{chexpert-plus} A publicly available dataset comprising 223,462 paired radiology reports and chest x-rays from 187,711 studies across 64,725 patients. We adopt the ReXrank test set, which follows the official CheXpert Plus test split and contains 200 studies. The CheXpert Plus training split is used for training \ours.

 \paragraph{ReXGradient} 
 \citep{rexgradient} A large-scale dataset curated by GradientHealth, consisting of a private testset of 10,000 studies from 7,004 patients across 67 clinical sites in the United States. The publicly released official training set, comprising 140,000 studies, is used for training \ours.

 \paragraph{IU-Xray} \citep{iuxray}
 A public dataset containing 7,470 radiology reports paired with corresponding frontal and lateral chest x-rays. We follow the ReXrank split and evaluate on the test set of 590 studies and train with the rest as the train set. To assess the zero-shot generalization capability of \ours on unseen data sources, we exclude the IU-Xray training set from the training corpus in the generalization study.

\paragraph{Proprietary Dataset (PD)}  
This proprietary dataset comprises 11,815 chest X-ray studies from inpatient and outpatient facilities across the United States. 
Each study includes corresponding radiology reports of frontal and lateral view images, without prior studies.
The dataset was used exclusively for evaluation, with no overlap with any training sources.

\bibliography{colm2025_conference}

\begin{thebibliography}{34}
\providecommand{\natexlab}[1]{#1}
\providecommand{\url}[1]{\texttt{#1}}
\expandafter\ifx\csname urlstyle\endcsname\relax
  \providecommand{\doi}[1]{doi: #1}\else
  \providecommand{\doi}{doi: \begingroup \urlstyle{rm}\Url}\fi

\bibitem[Bai et~al.(2025)Bai, Cai, Chen, Chen, Chen, Cheng, Deng, Ding, Gao,
  Ge, Ge, Guo, Huang, Huang, Huang, Hui, Jiang, Li, Li, Li, Li, Lin, Lin, Liu,
  Liu, Liu, Liu, Liu, Liu, Lu, Luo, Lv, Men, Meng, Ren, Ren, Song, Sun, Tang,
  Tu, Wan, Wang, Wang, Wang, Wang, Xie, Xu, Xu, Xu, Yang, Yang, Yang, Yang, Yu,
  Zhang, Zhang, Zhang, Zheng, Zhong, Zhou, Zhou, Zhou, Zhu, and
  Zhu]{bai2025qwen3vltechnicalreport}
Shuai Bai, Yuxuan Cai, Ruizhe Chen, Keqin Chen, Xionghui Chen, Zesen Cheng,
  Lianghao Deng, Wei Ding, Chang Gao, Chunjiang Ge, Wenbin Ge, Zhifang Guo,
  Qidong Huang, Jie Huang, Fei Huang, Binyuan Hui, Shutong Jiang, Zhaohai Li,
  Mingsheng Li, Mei Li, Kaixin Li, Zicheng Lin, Junyang Lin, Xuejing Liu,
  Jiawei Liu, Chenglong Liu, Yang Liu, Dayiheng Liu, Shixuan Liu, Dunjie Lu,
  Ruilin Luo, Chenxu Lv, Rui Men, Lingchen Meng, Xuancheng Ren, Xingzhang Ren,
  Sibo Song, Yuchong Sun, Jun Tang, Jianhong Tu, Jianqiang Wan, Peng Wang,
  Pengfei Wang, Qiuyue Wang, Yuxuan Wang, Tianbao Xie, Yiheng Xu, Haiyang Xu,
  Jin Xu, Zhibo Yang, Mingkun Yang, Jianxin Yang, An~Yang, Bowen Yu, Fei Zhang,
  Hang Zhang, Xi~Zhang, Bo~Zheng, Humen Zhong, Jingren Zhou, Fan Zhou, Jing
  Zhou, Yuanzhi Zhu, and Ke~Zhu.
\newblock Qwen3-vl technical report, 2025.
\newblock URL \url{https://arxiv.org/abs/2511.21631}.

\bibitem[Bannur et~al.(2023)Bannur, Hyland, Liu, Perez-Garcia, Ilse, Castro,
  Boecking, Sharma, Bouzid, Thieme, et~al.]{bannur2023learning}
Shruthi Bannur, Stephanie Hyland, Qianchu Liu, Fernando Perez-Garcia,
  Maximilian Ilse, Daniel~C Castro, Benedikt Boecking, Harshita Sharma, Kenza
  Bouzid, Anja Thieme, et~al.
\newblock Learning to exploit temporal structure for biomedical vision-language
  processing.
\newblock In \emph{Proceedings of the IEEE/CVF Conference on Computer Vision
  and Pattern Recognition}, pp.\  15016--15027, 2023.

\bibitem[Bannur et~al.(2024)Bannur, Bouzid, Castro, Schwaighofer, Thieme,
  Bond-Taylor, Ilse, Pérez-García, Salvatelli, Sharma, Meissen, Ranjit,
  Srivastav, Gong, Codella, Falck, Oktay, Lungren, Wetscherek, Alvarez-Valle,
  and Hyland]{maira-2}
Shruthi Bannur, Kenza Bouzid, Daniel~C. Castro, Anton Schwaighofer, Anja
  Thieme, Sam Bond-Taylor, Maximilian Ilse, Fernando Pérez-García, Valentina
  Salvatelli, Harshita Sharma, Felix Meissen, Mercy Ranjit, Shaury Srivastav,
  Julia Gong, Noel C.~F. Codella, Fabian Falck, Ozan Oktay, Matthew~P. Lungren,
  Maria~Teodora Wetscherek, Javier Alvarez-Valle, and Stephanie~L. Hyland.
\newblock Maira-2: Grounded radiology report generation, 2024.
\newblock URL \url{https://arxiv.org/abs/2406.04449}.

\bibitem[Chambon et~al.(2024)Chambon, Delbrouck, Sounack, Huang, Chen, Varma,
  Truong, Chuong, and Langlotz]{chexpert-plus}
Pierre Chambon, Jean-Benoit Delbrouck, Thomas Sounack, Shih-Cheng Huang,
  Zhihong Chen, Maya Varma, Steven~QH Truong, Chu~The Chuong, and Curtis~P.
  Langlotz.
\newblock Chexpert plus: Augmenting a large chest x-ray dataset with text
  radiology reports, patient demographics and additional image formats, 2024.
\newblock URL \url{https://arxiv.org/abs/2405.19538}.

\bibitem[Delbrouck et~al.(2025)Delbrouck, Xu, Moll, Thomas, Chen, Ostmeier,
  Azhar, Li, Johnston, Bluethgen, et~al.]{delbrouck2025automated}
Jean-Benoit Delbrouck, Justin Xu, Johannes Moll, Alois Thomas, Zhihong Chen,
  Sophie Ostmeier, Asfandyar Azhar, Kelvin~Zhenghao Li, Andrew Johnston,
  Christian Bluethgen, et~al.
\newblock Automated structured radiology report generation.
\newblock In \emph{Proceedings of the 63rd Annual Meeting of the Association
  for Computational Linguistics (Volume 1: Long Papers)}, pp.\  26813--26829,
  2025.

\bibitem[Demner-Fushman et~al.(2015)Demner-Fushman, Kohli, Rosenman, Shooshan,
  Rodriguez, Antani, Thoma, and McDonald]{iuxray}
Dina Demner-Fushman, Marc~D Kohli, Marc~B Rosenman, Sonya~E Shooshan, Laritza
  Rodriguez, Sameer Antani, George~R Thoma, and Clement~J McDonald.
\newblock Preparing a collection of radiology examinations for distribution and
  retrieval, 2015.

\bibitem[Devlin et~al.(2019)Devlin, Chang, Lee, and Toutanova]{devlin2019bert}
Jacob Devlin, Ming-Wei Chang, Kenton Lee, and Kristina Toutanova.
\newblock Bert: Pre-training of deep bidirectional transformers for language
  understanding.
\newblock In \emph{Proceedings of the 2019 conference of the North American
  chapter of the association for computational linguistics: human language
  technologies, volume 1 (long and short papers)}, pp.\  4171--4186, 2019.

\bibitem[Hartung et~al.(2020)Hartung, Bickle, Gaillard, and
  Kanne]{hartung2020create}
Michael~P Hartung, Ian~C Bickle, Frank Gaillard, and Jeffrey~P Kanne.
\newblock How to create a great radiology report.
\newblock \emph{Radiographics}, 40\penalty0 (6):\penalty0 1658--1670, 2020.

\bibitem[Huang et~al.(2023)Huang, Neill, Wittbrodt, Melnick, Klug, Thompson,
  Bailitz, Loftus, Malik, Phull, et~al.]{huang2023generative}
Jonathan Huang, Luke Neill, Matthew Wittbrodt, David Melnick, Matthew Klug,
  Michael Thompson, John Bailitz, Timothy Loftus, Sanjeev Malik, Amit Phull,
  et~al.
\newblock Generative artificial intelligence for chest radiograph
  interpretation in the emergency department.
\newblock \emph{JAMA network open}, 6\penalty0 (10):\penalty0
  e2336100--e2336100, 2023.

\bibitem[Jain et~al.(2021)Jain, Agrawal, Saporta, Truong, Duong, Bui, Chambon,
  Zhang, Lungren, Ng, et~al.]{jain1radgraph}
Saahil Jain, Ashwin Agrawal, Adriel Saporta, Steven Truong, Du~Nguyen Duong,
  Tan Bui, Pierre Chambon, Yuhao Zhang, Matthew~P Lungren, Andrew~Y Ng, et~al.
\newblock Radgraph: Extracting clinical entities and relations from radiology
  reports.
\newblock In \emph{Thirty-fifth Conference on Neural Information Processing
  Systems Datasets and Benchmarks Track (Round 1)}, 2021.

\bibitem[Johnson et~al.(2019)Johnson, Pollard, Berkowitz, Greenbaum, Lungren,
  Deng, Mark, and Horng]{mimic-cxr}
Alistair~EW Johnson, Tom~J Pollard, Seth~J Berkowitz, Nathaniel~R Greenbaum,
  Matthew~P Lungren, Chih-ying Deng, Roger~G Mark, and Steven Horng.
\newblock Mimic-cxr, a de-identified publicly available database of chest
  radiographs with free-text reports.
\newblock \emph{Scientific data}, 6\penalty0 (1):\penalty0 317, 2019.

\bibitem[Kahn~Jr et~al.(2009)Kahn~Jr, Langlotz, Burnside, Carrino, Channin,
  Hovsepian, and Rubin]{kahn2009toward}
Charles~E Kahn~Jr, Curtis~P Langlotz, Elizabeth~S Burnside, John~A Carrino,
  David~S Channin, David~M Hovsepian, and Daniel~L Rubin.
\newblock Toward best practices in radiology reporting.
\newblock \emph{Radiology}, 252\penalty0 (3):\penalty0 852--856, 2009.

\bibitem[Liu et~al.(2019)Liu, Hsu, McDermott, Boag, Weng, Szolovits, and
  Ghassemi]{liu2019clinically}
Guanxiong Liu, Tzu-Ming~Harry Hsu, Matthew McDermott, Willie Boag, Wei-Hung
  Weng, Peter Szolovits, and Marzyeh Ghassemi.
\newblock Clinically accurate chest x-ray report generation.
\newblock In \emph{Machine Learning for Healthcare Conference}, pp.\  249--269.
  PMLR, 2019.

\bibitem[Milam \& Koo(2023)Milam and Koo]{milam2023current}
ME~Milam and CW~Koo.
\newblock The current status and future of fda-approved artificial intelligence
  tools in chest radiology in the united states.
\newblock \emph{Clinical Radiology}, 78\penalty0 (2):\penalty0 115--122, 2023.

\bibitem[Nicolson et~al.(2023)Nicolson, Dowling, and
  Koopman]{nicolson_improving_2023}
Aaron Nicolson, Jason Dowling, and Bevan Koopman.
\newblock Improving chest {X}-ray report generation by leveraging warm
  starting.
\newblock \emph{Artificial Intelligence in Medicine}, 144:\penalty0 102633,
  2023.
\newblock ISSN 0933-3657.
\newblock \doi{https://doi.org/10.1016/j.artmed.2023.102633}.
\newblock URL
  \url{https://www.sciencedirect.com/science/article/pii/S0933365723001471}.

\bibitem[Papineni et~al.(2002)Papineni, Roukos, Ward, and
  Zhu]{papineni2002bleu}
Kishore Papineni, Salim Roukos, Todd Ward, and Wei-Jing Zhu.
\newblock Bleu: a method for automatic evaluation of machine translation.
\newblock In \emph{Proceedings of the 40th annual meeting of the Association
  for Computational Linguistics}, pp.\  311--318, 2002.

\bibitem[Sellergren et~al.(2025)Sellergren, Kazemzadeh, Jaroensri, Kiraly,
  Traverse, Kohlberger, Xu, Jamil, Hughes, Lau, Chen, Mahvar, Yatziv, Chen,
  Sterling, Baby, Baby, Lai, Schmidgall, Yang, Chen, Bjornsson, Reddy, Brush,
  Philbrick, Asiedu, Mezerreg, Hu, Yang, Tiwari, Jansen, Singh, Liu, Azizi,
  Kamath, Ferret, Pathak, Vieillard, Merhej, Perrin, Matejovicova, Ramé,
  Riviere, Rouillard, Mesnard, Cideron, bastien Grill, Ramos, Yvinec, Casbon,
  Buchatskaya, Alayrac, Lepikhin, Feinberg, Borgeaud, Andreev, Hardin, Dadashi,
  Hussenot, Joulin, Bachem, Matias, Chou, Hassidim, Goel, Farabet, Barral,
  Warkentin, Shlens, Fleet, Cotruta, Sanseviero, Martins, Kirk, Rao, Shetty,
  Steiner, Kirmizibayrak, Pilgrim, Golden, and Yang]{medgemma}
Andrew Sellergren, Sahar Kazemzadeh, Tiam Jaroensri, Atilla Kiraly, Madeleine
  Traverse, Timo Kohlberger, Shawn Xu, Fayaz Jamil, Cían Hughes, Charles Lau,
  Justin Chen, Fereshteh Mahvar, Liron Yatziv, Tiffany Chen, Bram Sterling,
  Stefanie~Anna Baby, Susanna~Maria Baby, Jeremy Lai, Samuel Schmidgall,
  Lu~Yang, Kejia Chen, Per Bjornsson, Shashir Reddy, Ryan Brush, Kenneth
  Philbrick, Mercy Asiedu, Ines Mezerreg, Howard Hu, Howard Yang, Richa Tiwari,
  Sunny Jansen, Preeti Singh, Yun Liu, Shekoofeh Azizi, Aishwarya Kamath, Johan
  Ferret, Shreya Pathak, Nino Vieillard, Ramona Merhej, Sarah Perrin, Tatiana
  Matejovicova, Alexandre Ramé, Morgane Riviere, Louis Rouillard, Thomas
  Mesnard, Geoffrey Cideron, Jean bastien Grill, Sabela Ramos, Edouard Yvinec,
  Michelle Casbon, Elena Buchatskaya, Jean-Baptiste Alayrac, Dmitry Lepikhin,
  Vlad Feinberg, Sebastian Borgeaud, Alek Andreev, Cassidy Hardin, Robert
  Dadashi, Léonard Hussenot, Armand Joulin, Olivier Bachem, Yossi Matias,
  Katherine Chou, Avinatan Hassidim, Kavi Goel, Clement Farabet, Joelle Barral,
  Tris Warkentin, Jonathon Shlens, David Fleet, Victor Cotruta, Omar
  Sanseviero, Gus Martins, Phoebe Kirk, Anand Rao, Shravya Shetty, David~F.
  Steiner, Can Kirmizibayrak, Rory Pilgrim, Daniel Golden, and Lin Yang.
\newblock Medgemma technical report, 2025.
\newblock URL \url{https://arxiv.org/abs/2507.05201}.

\bibitem[Serra et~al.(2023)Serra, Wang, Deligianni, Dalton, and
  O'Neil]{serra2023controllable}
Francesco~Dalla Serra, Chaoyang Wang, Fani Deligianni, Jeffrey Dalton, and
  Alison~Q O'Neil.
\newblock Controllable chest x-ray report generation from longitudinal
  representations.
\newblock \emph{arXiv preprint arXiv:2310.05881}, 2023.

\bibitem[Shao et~al.(2025)Shao, Asai, Shen, Ivison, Kishore, Zhuo, Zhao, Park,
  Finlayson, Sontag, et~al.]{shao2025dr}
Rulin Shao, Akari Asai, Shannon~Zejiang Shen, Hamish Ivison, Varsha Kishore,
  Jingming Zhuo, Xinran Zhao, Molly Park, Samuel~G Finlayson, David Sontag,
  et~al.
\newblock Dr tulu: Reinforcement learning with evolving rubrics for deep
  research.
\newblock \emph{arXiv preprint arXiv:2511.19399}, 2025.

\bibitem[Shao et~al.(2024)Shao, Wang, Zhu, Xu, Song, Bi, Zhang, Zhang, Li,
  et~al.]{shao2024deepseekmath}
Zhihong Shao, Peiyi Wang, Qihao Zhu, Runxin Xu, Junxiao Song, Xiao Bi, Haowei
  Zhang, Mingchuan Zhang, YK~Li, et~al.
\newblock Deepseekmath: Pushing the limits of mathematical reasoning in open
  language models.
\newblock \emph{arXiv preprint arXiv:2402.03300}, 2024.

\bibitem[Smit et~al.(2020)Smit, Jain, Rajpurkar, Pareek, Ng, and
  Lungren]{chexbert}
Akshay Smit, Saahil Jain, Pranav Rajpurkar, Anuj Pareek, Andrew~Y. Ng, and
  Matthew~P. Lungren.
\newblock Chexbert: Combining automatic labelers and expert annotations for
  accurate radiology report labeling using bert, 2020.
\newblock URL \url{https://arxiv.org/abs/2004.09167}.

\bibitem[Tanida et~al.(2023)Tanida, Müller, Kaissis, and
  Rueckert]{tanida2023interactive}
Tim Tanida, Philip Müller, Georgios Kaissis, and Daniel Rueckert.
\newblock Interactive and explainable region-guided radiology report
  generation.
\newblock In \emph{CVPR}, 2023.

\bibitem[Yu et~al.(2022)Yu, Mohajer, and Eng]{yu2022external}
Alice~C Yu, Bahram Mohajer, and John Eng.
\newblock External validation of deep learning algorithms for radiologic
  diagnosis: a systematic review.
\newblock \emph{Radiology: Artificial Intelligence}, 4\penalty0 (3):\penalty0
  e210064, 2022.

\bibitem[Yu et~al.(2023{\natexlab{a}})Yu, Endo, Krishnan, Pan, Tsai, Reis,
  Fonseca, Lee, Abad, Ng, et~al.]{radcliq}
Feiyang Yu, Mark Endo, Rayan Krishnan, Ian Pan, Andy Tsai, Eduardo~Pontes Reis,
  Eduardo Kaiser Ururahy~Nunes Fonseca, Henrique Min~Ho Lee, Zahra
  Shakeri~Hossein Abad, Andrew~Y Ng, et~al.
\newblock Evaluating progress in automatic chest x-ray radiology report
  generation.
\newblock \emph{Patterns}, 4\penalty0 (9), 2023{\natexlab{a}}.

\bibitem[Yu et~al.(2023{\natexlab{b}})Yu, Endo, Krishnan, Pan, Tsai, Reis,
  Fonseca, Lee, Abad, Ng, et~al.]{yu2023evaluating}
Feiyang Yu, Mark Endo, Rayan Krishnan, Ian Pan, Andy Tsai, Eduardo~Pontes Reis,
  Eduardo Kaiser Ururahy~Nunes Fonseca, Henrique Min~Ho Lee, Zahra
  Shakeri~Hossein Abad, Andrew~Y Ng, et~al.
\newblock Evaluating progress in automatic chest x-ray radiology report
  generation.
\newblock \emph{Patterns}, 4\penalty0 (9), 2023{\natexlab{b}}.

\bibitem[Yu et~al.(2025)Yu, Zhang, Zhu, Yuan, Zuo, Yue, Dai, Fan, Liu, Liu,
  et~al.]{yu2025dapo}
Qiying Yu, Zheng Zhang, Ruofei Zhu, Yufeng Yuan, Xiaochen Zuo, Yu~Yue, Weinan
  Dai, Tiantian Fan, Gaohong Liu, Lingjun Liu, et~al.
\newblock Dapo: An open-source llm reinforcement learning system at scale.
\newblock \emph{arXiv preprint arXiv:2503.14476}, 2025.

\bibitem[Zambrano~Chaves et~al.(2025{\natexlab{a}})Zambrano~Chaves, Huang, Xu,
  Xu, Usuyama, Zhang, Wang, Xie, Khademi, Yang, et~al.]{llava-rad}
Juan~Manuel Zambrano~Chaves, Shih-Cheng Huang, Yanbo Xu, Hanwen Xu, Naoto
  Usuyama, Sheng Zhang, Fei Wang, Yujia Xie, Mahmoud Khademi, Ziyi Yang, et~al.
\newblock A clinically accessible small multimodal radiology model and
  evaluation metric for chest x-ray findings.
\newblock \emph{Nature Communications}, 16\penalty0 (1):\penalty0 3108,
  2025{\natexlab{a}}.

\bibitem[Zambrano~Chaves et~al.(2025{\natexlab{b}})Zambrano~Chaves, Huang, Xu,
  Xu, Usuyama, Zhang, Wang, Xie, Khademi, Yang, et~al.]{zambrano2025clinically}
Juan~Manuel Zambrano~Chaves, Shih-Cheng Huang, Yanbo Xu, Hanwen Xu, Naoto
  Usuyama, Sheng Zhang, Fei Wang, Yujia Xie, Mahmoud Khademi, Ziyi Yang, et~al.
\newblock A clinically accessible small multimodal radiology model and
  evaluation metric for chest x-ray findings.
\newblock \emph{Nature Communications}, 16\penalty0 (1):\penalty0 3108,
  2025{\natexlab{b}}.

\bibitem[Zhang et~al.(2019)Zhang, Kishore, Wu, Weinberger, and
  Artzi]{zhangbertscore}
Tianyi Zhang, Varsha Kishore, Felix Wu, Kilian~Q Weinberger, and Yoav Artzi.
\newblock Bertscore: Evaluating text generation with bert.
\newblock In \emph{International Conference on Learning Representations}, 2019.

\bibitem[Zhang et~al.(2024)Zhang, Zhou, Yang, Banerjee, Acosta, Miller, Huang,
  and Rajpurkar]{rexrank}
Xiaoman Zhang, Hong-Yu Zhou, Xiaoli Yang, Oishi Banerjee, Juli{\'a}n~N Acosta,
  Josh Miller, Ouwen Huang, and Pranav Rajpurkar.
\newblock Rexrank: A public leaderboard for ai-powered radiology report
  generation.
\newblock \emph{arXiv preprint arXiv:2411.15122}, 2024.

\bibitem[Zhang et~al.(2025)Zhang, Acosta, Miller, Huang, and
  Rajpurkar]{rexgradient}
Xiaoman Zhang, Julián~N. Acosta, Josh Miller, Ouwen Huang, and Pranav
  Rajpurkar.
\newblock Rexgradient-160k: A large-scale publicly available dataset of chest
  radiographs with free-text reports, 2025.
\newblock URL \url{https://arxiv.org/abs/2505.00228}.

\bibitem[Zhao et~al.(2024)Zhao, Wu, Zhang, Zhang, Wang, and
  Xie]{zhao2024ratescore}
Weike Zhao, Chaoyi Wu, Xiaoman Zhang, Ya~Zhang, Yanfeng Wang, and Weidi Xie.
\newblock Ratescore: A metric for radiology report generation.
\newblock In \emph{Proceedings of the 2024 Conference on Empirical Methods in
  Natural Language Processing}, pp.\  15004--15019, 2024.

\bibitem[Zhou et~al.(2025)Zhou, Acosta, Adithan, Datta, Topol, and
  Rajpurkar]{medversa}
Hong-Yu Zhou, Julián~Nicolás Acosta, Subathra Adithan, Suvrankar Datta,
  Eric~J. Topol, and Pranav Rajpurkar.
\newblock Medversa: A generalist foundation model for medical image
  interpretation, 2025.
\newblock URL \url{https://arxiv.org/abs/2405.07988}.

\bibitem[Zhu et~al.(2023)Zhu, Mathai, Mukherjee, Peng, Summers, and
  Lu]{zhu2023utilizing}
Qingqing Zhu, Tejas~Sudharshan Mathai, Pritam Mukherjee, Yifan Peng, Ronald~M
  Summers, and Zhiyong Lu.
\newblock Utilizing longitudinal chest x-rays and reports to pre-fill radiology
  reports.
\newblock In \emph{International Conference on Medical Image Computing and
  Computer-Assisted Intervention}, pp.\  189--198. Springer, 2023.

\end{thebibliography}
\bibliographystyle{colm2025_conference}

\appendix
\section{Supplementary Information}

\subsection{Comparing SFT and RL\label{sec:sft_vs_rl}}

To identify the most effective training strategy for radiology report generation, we compare three approaches on the MIMIC dataset: SFT alone, RL alone, and the combined SFT~+~RL pipeline. All SFT and RL runs are trained for 3 epochs, while the SFT~+~RL configuration applies an additional 2 epochs of RL starting from the SFT checkpoint. As shown in \Cref{Table: sft vs rl}, RL alone outperforms SFT on BERTScore and SembScore, but lags significantly behind on RadGraph-F1. This is likely because RadGraph is more sensitive to specific wording, phrasing, and lexical distributions, which SFT captures more directly through next-token prediction. Consequently, SFT slightly outperforms RL on the overall RadCliQ score. The combined \emph{SFT~+~RL} setup yields the strongest results across all metrics. SFT first teaches the model the output format and dataset-specific lexical structure, and RL subsequently refines this foundation by optimizing more semantically and clinically aligned rewards, leading to the best synergy and overall performance.

\begin{table}[htbp]
\centering
\caption{Comparison of SFT, RL, and SFT~+~RL on findings + impression generation on MIMIC. The combined SFT~+~RL strategy achieves the best performance.\label{Table: sft vs rl}}
\begin{tabular}{lcccc}
\toprule
\textbf{Method} & \textbf{BERTScore} & \textbf{SembScore} & \textbf{RadGraph-F1} & \textbf{1/RadCliQ} \\
\midrule
Baseline & 0.293 & 0.242 & 0.102 & 0.625 \\
MIMIC SFT                & 0.421 & 0.426 & 0.236 & 0.962 \\
MIMIC RL                 & 0.437 & 0.435 & 0.198 & 0.950 \\
MIMIC SFT + MIMIC RL     & \textbf{0.449} & \textbf{0.478} & \textbf{0.267} & \textbf{1.110} \\
\bottomrule
\end{tabular}
\end{table}

\subsection{Methodological Ablations for UniRG\label{sec:unirg_method_ablation}}

The following ablations clarify why \ours is not equivalent to a generic SFT~+~RL or LLM-as-a-judge setup. Our contribution lies in the task-specific RL recipe for balancing multiple clinically grounded objectives, rather than in using RL alone.

\begin{table}[!ht]
\centering
\begin{tabular}{lc}
\toprule
\textbf{Setup} & \textbf{1/RadCliQ} \\
\midrule
\ours & 1.23 \\
LLM-as-a-judge RL & 0.77 \\
\bottomrule
\end{tabular}
\caption{Comparison on the MIMIC test set between \ours and an LLM-as-a-judge RL setup based on Dr.~Tulu~\citep{shao2025dr}. Directly applying an LLM-as-a-judge reward underperforms \ours substantially on radiology report generation. \label{tab:llm_judge_comparison}}
\end{table}

\begin{table}[!ht]
\centering
\begin{tabular}{l c c c c}
\hline
\textbf{Setup} & \textbf{Steps} & \textbf{Reward} & \textbf{KL Reg.} & \textbf{Performance Drop} \\
\hline
\ours & two-step & composite & yes & - \\
ablation A & one-step & single & no & 11.4\% \\
ablation B & one-step & composite & no & 4.9\% \\
ablation C & two-step & composite & no & 2.4\% \\
\hline
\end{tabular}
\caption{Comparison of RL design choices on the MIMIC test set. Performance drop is computed relative to \ours based on 1/RadCliQ. The full two-step composite-reward design with KL regularization yields the best performance. \label{tab:unirg_ablation_recipe}}
\end{table}

These results show that the strongest performance comes from the full \ours recipe: staged optimization improves the balance across objectives, the composite reward is better than optimizing a single signal, and KL regularization is necessary to stabilize the second-stage error-reduction step.

\subsection{Prompt Templates\label{appendix:prompt}}
Below are training and inference prompt templates for findings + impression generation and findings generation. \{input\} is the placeholder for the input images and the context from the indication and comparison sections.

\begin{tcolorbox}[colback=gray!5,colframe=gray!80,
  listing only,breakable,arc=2mm,title={Findings + Impression Generation Prompt}]

This is a radiology report generation task. Here is the context: \{input\}
Given the image and the context, provide the report in the following format:
Findings: [write the findings]
Impression: [write the impression]
Now write the report in the format above.

\end{tcolorbox}

\begin{tcolorbox}[colback=gray!5,colframe=gray!80,
  listing only,breakable,arc=2mm,title={Findings Generation Prompt}]

This is a radiology report generation task. Here is the context: \{input\}
Given the image and the context, provide the findings in the following format:
Findings: [write the findings]
Now write the report in the format above.

\end{tcolorbox}

\subsection{Statistical Significance Testing}
\label{sec:stats-test}

The tables below report statistical significance tests for all results presented in the paper, except for \Cref{fig:overview-of-unirg-cxr,fig:sota-report-gen}, which are sourced from the ReXrank leaderboard with undisclosed evaluation data. Results in \Cref{tab:3a,tab:3b,tab:3c,tab:4a} demonstrate that the performance gains of \ours are statistically significant. Additionally, results in \Cref{tab:4d} show that there is no statistically significant difference across demographic subgroups, indicating consistent performance across the board by \ours.

\begin{table}[!ht]
\centering
\begin{tabular}{cccrrrc}
\toprule
\textbf{metric} & \textbf{model\_a} & \textbf{model\_b} & \textbf{mean\_a} & \textbf{mean\_b} & \textbf{diff} & \textbf{p\_value} \\ \midrule
\multirow{4}{*}{1/RadCliQ-v1} & \multirow{4}{*}{UniRG-CXR} & Maira-2 & 1.31 & 1.07 & 0.25 & $p<0.0001$ \\
 &  & GPT-5 & 1.31 & 0.86 & 0.45 & $p<0.0001$ \\
 &  & MedGemma & 1.31 & 0.84 & 0.48 & $p<0.0001$ \\
 &  & GPT-4o & 1.31 & 0.82 & 0.49 & $p<0.0001$ \\ \midrule
\multirow{4}{*}{BLEU} & \multirow{4}{*}{UniRG-CXR} & Maira-2 & 26.59 & 24.02 & 2.57 & $p<0.0001$ \\
 &  & GPT-5 & 26.59 & 14.26 & 12.32 & $p<0.0001$ \\
 &  & MedGemma & 26.59 & 17.16 & 9.43 & $p<0.0001$ \\
 &  & GPT-4o & 26.59 & 18.72 & 7.87 & $p<0.0001$ \\ \midrule
\multirow{4}{*}{BERTScore} & \multirow{4}{*}{UniRG-CXR} & Maira-2 & 50.54 & 47.13 & 3.41 & $p<0.0001$ \\
 &  & GPT-5 & 50.54 & 39.76 & 10.77 & $p<0.0001$ \\
 &  & MedGemma & 50.54 & 40.33 & 10.21 & $p<0.0001$ \\
 &  & GPT-4o & 50.54 & 39.96 & 10.58 & $p<0.0001$ \\ \midrule
\multirow{4}{*}{SembScore} & \multirow{4}{*}{UniRG-CXR} & Maira-2 & 50.85 & 44.29 & 6.56 & $p<0.0001$ \\
 &  & GPT-5 & 50.85 & 39.65 & 11.20 & $p<0.0001$ \\
 &  & MedGemma & 50.85 & 36.99 & 13.87 & $p<0.0001$ \\
 &  & GPT-4o & 50.85 & 35.71 & 15.14 & $p<0.0001$ \\ \midrule
\multirow{4}{*}{RadGraph} & \multirow{4}{*}{UniRG-CXR} & Maira-2 & 28.03 & 23.60 & 4.43 & $p<0.0001$ \\
 &  & GPT-5 & 28.03 & 19.46 & 8.57 & $p<0.0001$ \\
 &  & MedGemma & 28.03 & 17.86 & 10.17 & $p<0.0001$ \\
 &  & GPT-4o & 28.03 & 17.90 & 10.13 & $p<0.0001$ \\ \bottomrule
\end{tabular}
\caption{
Statistical significance of pairwise model comparisons in the longitudinal report generation setup (Figure 3a).
\ours is compared against each baseline using a bootstrap z-test (1,200 bootstrap resamples). For each comparison, the z-statistic is
computed as mean(diffs) / std(diffs) from the paired bootstrap distributions,
and the p-value is derived from the standard normal distribution (two-sided).
 mean\_a:      Mean of bootstrap sample means for model\_a.
 mean\_b:      Mean of bootstrap sample means for model\_b.
 diff:        Difference (mean\_a - mean\_b). Positive values indicate model\_a
                 outperforms model\_b for all metrics.
 p\_value:     Two-sided p-value from the bootstrap z-test.
}
 \label{tab:3a}
\end{table}

\begin{table}[!ht]
\centering
\begin{tabular}{@{}ccccrrrc@{}}
\toprule
\textbf{encounter} & \textbf{metric} &  \textbf{model\_a} &\textbf{model\_b} & \textbf{mean\_a} & \textbf{mean\_b} & \textbf{diff} & \textbf{p\_value} \\ \midrule
\multirow{4}{*}{1st} & \multirow{4}{*}{RadCliQ-v1}  & \multirow{4}{*}{\ours}& Maira-2 & 1.23 & 1.00 & 0.23 & $p<0.0001$ \\
 &  & & MedGemma & 1.23 & 0.74 & 0.49 & $p<0.0001$ \\
 &  & & GPT-4o & 1.23 & 0.71 & 0.53 & $p<0.0001$ \\
 &  & & GPT-5 & 1.23 & 0.79 & 0.44 & $p<0.0001$ \\ \midrule
\multirow{5}{*}{2nd} & \multirow{5}{*}{RadCliQ-v1} & \multirow{4}{*}{\ours} & Maira-2 & 1.36 & 1.12 & 0.23 & $p<0.0001$ \\
 &  & & MedGemma & 1.36 & 0.85 & 0.50 & $p<0.0001$ \\
 &  && GPT-4o & 1.36 & 0.87 & 0.49 & $p<0.0001$ \\
 &  && GPT-5 & 1.36 & 0.90 & 0.46 & $p<0.0001$ \\
 &  && Copy Prior & 1.36 & 0.81 & 0.54 & $p<0.0001$ \\ \midrule
\multirow{5}{*}{3rd} & \multirow{5}{*}{RadCliQ-v1} & \multirow{4}{*}{\ours} & Maira-2 & 1.34 & 1.04 & 0.29 & $p<0.0001$ \\
 &  && MedGemma & 1.34 & 0.90 & 0.43 & $p<0.0001$ \\
 &  && GPT-4o & 1.34 & 0.90 & 0.43 & $p<0.0001$ \\
 &  && GPT-5 & 1.34 & 0.89 & 0.44 & $p<0.0001$ \\
 &  && Copy Prior & 1.34 & 0.87 & 0.47 & $p<0.0001$ \\ \midrule
\multirow{5}{*}{4th} & \multirow{5}{*}{RadCliQ-v1} & \multirow{4}{*}{\ours} & Maira-2 & 1.35 & 1.15 & 0.19 & $p<0.0001$ \\
 &  && MedGemma & 1.35 & 0.92 & 0.43 & $p<0.0001$ \\
 &  && GPT-4o & 1.35 & 0.92 & 0.43 & $p<0.0001$ \\
 &  && GPT-5 & 1.35 & 0.93 & 0.42 & $p<0.0001$ \\
 &  && Copy Prior & 1.35 & 0.87 & 0.47 & $p<0.0001$ \\ \midrule
\multirow{5}{*}{5th+} & \multirow{5}{*}{RadCliQ-v1} & \multirow{4}{*}{\ours} & Maira-2 & 1.39 & 1.10 & 0.29 & $p<0.0001$ \\
 &  && MedGemma & 1.39 & 0.94 & 0.46 & $p<0.0001$ \\
 &  && GPT-4o & 1.39 & 0.94 & 0.46 & $p<0.0001$ \\
 &  && GPT-5 & 1.39 & 0.92 & 0.47 & $p<0.0001$ \\
 &  && Copy Prior & 1.39 & 0.90 & 0.49 & $p<0.0001$ \\ \bottomrule
\end{tabular}
\caption{
Statistical significance of pairwise model comparisons by encounter number (Figure 3b).
\ours is compared against each baseline using a bootstrap z-test (1,200 bootstrap resamples).
The z-statistic is computed as mean(diffs) / std(diffs) from the paired bootstrap distributions,
and the p-value is derived from the standard normal distribution (two-sided).
Copy Prior is excluded for the 1st encounter (no prior report available).
 mean\_a:      Mean of bootstrap sample means for model\_a.
 mean\_b:      Mean of bootstrap sample means for model\_b.
 diff:        Difference (mean\_a - mean\_b). Positive values indicate model\_a
                 outperforms model\_b for all metrics.
 p\_value:     Two-sided p-value from the bootstrap z-test.
 \label{tab:3b}
}
\end{table}

\begin{table}[!ht]
\centering
\begin{tabular}{@{}ccccrrrc@{}}
\toprule
\textbf{subgroup} & \textbf{metric} & \textbf{model\_a} & \textbf{model\_b} & \textbf{mean\_a} & \textbf{mean\_b} & \textbf{diff} & \textbf{p\_value} \\ \midrule
\multirow{3}{*}{First Study} & \multirow{3}{*}{RadCliQ-v1}& \multirow{3}{*}{\ours}  & Maira-2 & 1.23 & 1.00 & 0.23 & $p<0.0001$ \\
&  &  & GPT-5 & 1.23 & 0.79 & 0.44 & $p<0.0001$ \\
&  &  & MedGemma & 1.23 & 0.74 & 0.49 & $p<0.0001$ \\ \midrule
\multirow{4}{*}{New\,Development} & \multirow{4}{*}{RadCliQ-v1} & \multirow{4}{*}{\ours} & Maira-2 & 1.35 & 1.09 & 0.26 & $p<0.0001$ \\
&  &  & GPT-5 & 1.35 & 0.86 & 0.49 & $p<0.0001$ \\
&  &  & MedGemma & 1.35 & 0.83 & 0.52 & $p<0.0001$ \\
&  &  & Copy Prior & 1.35 & 0.78 & 0.57 & $p<0.0001$ \\ \midrule
\multirow{4}{*}{No Change} & \multirow{4}{*}{RadCliQ-v1} & \multirow{4}{*}{\ours} & Maira-2 & 1.48 & 1.20 & 0.28 & $p<0.0001$ \\
&  &  & GPT-5 & 1.48 & 0.98 & 0.51 & $p<0.0001$ \\
&  &  & MedGemma & 1.48 & 0.96 & 0.52 & $p<0.0001$ \\
&  &  & Copy Prior & 1.48 & 0.93 & 0.55 & $p<0.0001$ \\ \midrule
\multirow{4}{*}{Progression} & \multirow{4}{*}{RadCliQ-v1} & \multirow{4}{*}{\ours}  & Maira-2 & 1.26 & 0.99 & 0.27 & $p<0.0001$ \\
&  &  & GPT-5 & 1.26 & 0.85 & 0.41 & $p<0.0001$ \\
&  &  & MedGemma & 1.26 & 0.85 & 0.41 & $p<0.0001$ \\
&  &  & Copy Prior & 1.26 & 0.82 & 0.45 & $p<0.0001$ \\ \midrule
\multirow{4}{*}{Regression} & \multirow{4}{*}{RadCliQ-v1} & \multirow{4}{*}{\ours} & Maira-2 & 1.19 & 0.99 & 0.20 & $p<0.0001$ \\
&  &  & GPT-5 & 1.19 & 0.85 & 0.34 & $p<0.0001$ \\
&  &  & MedGemma & 1.19 & 0.86 & 0.34 & $p<0.0001$ \\
&  &  & Copy Prior & 1.19 & 0.82 & 0.37 & $p<0.0001$ \\ \bottomrule
\end{tabular}
\caption{
Statistical significance of pairwise model comparisons by clinical label (Figure 3c).
\ours is compared against each baseline using a bootstrap z-test (1,200 bootstrap resamples).
The z-statistic is computed as mean(diffs) / std(diffs) from the paired bootstrap distributions,
and the p-value is derived from the standard normal distribution (two-sided).
Copy Prior is excluded for the first encounter label (no prior report available).
 mean\_a:      Mean of bootstrap sample means for model\_a.
 mean\_b:      Mean of bootstrap sample means for model\_b.
 diff:        Difference (mean\_a - mean\_b). Positive values indicate model\_a
                 outperforms model\_b for all metrics.
 p\_value:     Two-sided p-value from the bootstrap z-test.
 \label{tab:3c}
}
\end{table}

\begin{table}[!ht]
\centering
\begin{tabular}{@{}ccccrrrc@{}}
\toprule
\textbf{dataset} & \textbf{metric} & \textbf{model\_a} & \textbf{model\_b} & \textbf{mean\_a} & \textbf{mean\_b} & \textbf{diff} & \textbf{p\_value} \\ \midrule
\multirow{20}{*}{IU-Xray} & \multirow{4}{*}{RadCliQ-v1} & \multirow{4}{*}{\ours} & GPT-4o & 2.01 & 1.09 & 0.91 & $p<0.0001$ \\
&  &  & GPT-5 & 2.01 & 1.12 & 0.89 & $p<0.0001$ \\
&  &  & MedGemma & 2.01 & 1.48 & 0.53 & $p<0.0001$ \\
&  &  & MedVersa & 2.01 & 1.48 & 0.53 & $p<0.0001$ \\ \cmidrule(l){2-8} 
 & \multirow{4}{*}{BLEU} & \multirow{4}{*}{\ours} & GPT-4o & 26.75 & 16.83 & 9.91 & $p<0.0001$ \\
&  &  & GPT-5 & 26.75 & 18.72 & 8.03 & $p<0.0001$ \\
&  &  & MedGemma & 26.75 & 23.56 & 3.19 & $p<0.0001$ \\
&  &  & MedVersa & 26.75 & 20.77 & 5.98 & $p<0.0001$ \\ \cmidrule(l){2-8} 
 & \multirow{4}{*}{BERTScore} & \multirow{4}{*}{\ours} & GPT-4o & 56.69 & 43.46 & 13.23 & $p<0.0001$ \\
&  &  & GPT-5 & 56.69 & 44.34 & 12.35 & $p<0.0001$ \\
&  &  & MedGemma & 56.69 & 51.98 & 4.71 & $p<0.0001$ \\
&  &  & MedVersa & 56.69 & 52.40 & 4.28 & $p<0.0001$ \\ \cmidrule(l){2-8} 
 & \multirow{4}{*}{SembScore} & \multirow{4}{*}{\ours} & GPT-4o & 66.02 & 55.14 & 10.88 & $p<0.0001$ \\
&  &  & GPT-5 & 66.02 & 54.07 & 11.95 & $p<0.0001$ \\
&  &  & MedGemma & 66.02 & 61.21 & 4.81 & $p<0.0001$ \\
&  &  & MedVersa & 66.02 & 61.39 & 4.63 & $p<0.0001$ \\ \cmidrule(l){2-8} 
 & \multirow{4}{*}{RadGraph} & \multirow{4}{*}{\ours} & GPT-4o & 28.84 & 21.67 & 7.17 & $p<0.0001$ \\
&  &  & GPT-5 & 28.84 & 23.16 & 5.67 & $p<0.0001$ \\
&  &  & MedGemma & 28.84 & 24.49 & 4.34 & $p<0.0001$ \\
&  &  & MedVersa & 28.84 & 23.78 & 5.06 & $p<0.0001$ \\ \midrule
\multirow{20}{*}{PD} & \multirow{4}{*}{RadCliQ-v1} & \multirow{4}{*}{\ours} & GPT-4o & 1.11 & 0.76 & 0.34 & $p<0.0001$ \\
&  &  & GPT-5 & 1.11 & 0.71 & 0.40 & $p<0.0001$ \\
&  &  & MedGemma & 1.11 & 0.92 & 0.19 & $p<0.0001$ \\
&  &  & MedVersa & 1.11 & 0.89 & 0.21 & $p<0.0001$ \\ \cmidrule(l){2-8} 
 & \multirow{4}{*}{BLEU} & \multirow{4}{*}{\ours} & GPT-4o & 20.42 & 12.28 & 8.14 & $p<0.0001$ \\
&  &  & GPT-5 & 20.42 & 11.16 & 9.26 & $p<0.0001$ \\
&  &  & MedGemma & 20.42 & 16.69 & 3.73 & $p<0.0001$ \\
&  &  & MedVersa & 20.42 & 15.95 & 4.47 & $p<0.0001$ \\ \cmidrule(l){2-8} 
 & \multirow{4}{*}{BERTScore} & \multirow{4}{*}{\ours} & GPT-4o & 47.40 & 33.48 & 13.92 & $p<0.0001$ \\
&  &  & GPT-5 & 47.40 & 30.69 & 16.71 & $p<0.0001$ \\
&  &  & MedGemma & 47.40 & 43.52 & 3.88 & $p<0.0001$ \\
&  &  & MedVersa & 47.40 & 41.68 & 5.72 & $p<0.0001$ \\ \cmidrule(l){2-8} 
 & \multirow{4}{*}{SembScore} & \multirow{4}{*}{\ours} & GPT-4o & 52.38 & 41.88 & 10.50 & $p<0.0001$ \\
&  &  & GPT-5 & 52.38 & 37.48 & 14.89 & $p<0.0001$ \\
&  &  & MedGemma & 52.38 & 45.54 & 6.84 & $p<0.0001$ \\
&  &  & MedVersa & 52.38 & 45.11 & 7.26 & $p<0.0001$ \\ \cmidrule(l){2-8} 
 & \multirow{4}{*}{RadGraph} & \multirow{4}{*}{\ours} & GPT-4o & 19.47 & 13.47 & 5.99 & $p<0.0001$ \\
&  &  & GPT-5 & 19.47 & 12.40 & 7.06 & $p<0.0001$ \\
&  &  & MedGemma & 19.47 & 15.42 & 4.05 & $p<0.0001$ \\
&  &  & MedVersa & 19.47 & 15.42 & 4.05 & $p<0.0001$ \\ \bottomrule
\end{tabular}
\caption{
Statistical significance of pairwise model comparisons on zero-shot transfer datasets (Figure 4a).
\ours is compared against each baseline using a bootstrap z-test (1,200 bootstrap resamples).
The z-statistic is computed as mean(diffs) / std(diffs) from the paired bootstrap distributions,
and the p-value is derived from the standard normal distribution (two-sided).
 mean\_a:      Mean of bootstrap sample means for model\_a.
 mean\_b:      Mean of bootstrap sample means for model\_b.
 diff:        Difference (mean\_a - mean\_b). Positive values indicate model\_a
                 outperforms model\_b for all metrics.
 p\_value:     Two-sided p-value from the bootstrap z-test.
 \label{tab:4a}
}
\end{table}

\begin{table}[!ht]
\centering
\begin{tabular}{@{}cccrrrl@{}}
\toprule
\textbf{category} & \textbf{subgroup} & \textbf{comparison} & \textbf{mean\_a} & \textbf{mean\_b} & \textbf{diff} & \textbf{p\_value} \\ \midrule
\multirow{5}{*}{Gender} & All & Female vs Male & 1.01 & 1.02 & -0.01 & $p \geq 0.05$ \\
 & $<60$ & Female vs Male & 0.96 & 1.04 & -0.09 & $p \geq 0.05$ \\
 & $\geq60$ & Female vs Male & 1.06 & 1.02 & 0.04 & $p \geq 0.05$ \\
 & White & Female vs Male & 1.13 & 1.03 & 0.11 & $p \geq 0.05$ \\
 & Non-white & Female vs Male & 0.92 & 1.04 & -0.12 & $p \geq 0.05$ \\ \midrule
\multirow{3}{*}{Age} & All & $<60$ vs $\geq60$ & 0.99 & 1.03 & -0.04 & $p \geq 0.05$ \\
 & Female & $<60$ vs $\geq60$ & 0.96 & 1.06 & -0.1 & $p \geq 0.05$ \\
 & White & $<60$ vs $\geq60$ & 1.11 & 1.04 & 0.08 & $p \geq 0.05$ \\ \midrule
\multirow{4}{*}{Race} & All & White vs Non-white & 1.06 & 0.96 & 0.1 & $p \geq 0.05$ \\
 & Female & White vs Non-white & 1.13 & 0.92 & 0.22 & $p \geq 0.05$ \\
 & Male & White vs Non-white & 1.03 & 1.04 & -0.01 & $p \geq 0.05$ \\
 & $\geq60$ & White vs Non-white & 1.04 & 1.03 & 0.01 & $p \geq 0.05$ \\ 
\bottomrule
\end{tabular}
\caption{
Statistical significance of demographic fairness comparisons (Figure 4d).
\ours 1/RadCliQ-v1 scores are compared across demographic subgroups using a
bootstrap z-test (1,200 bootstrap resamples). No comparison is statistically
significant (p $<$ 0.05), indicating equitable performance across demographics.
mean\_a and mean\_b correspond to the two groups in each comparison.
p\_value: two-sided p-value.
 \label{tab:4d}
}
\end{table}

\clearpage

\subsection{Inter-annotator agreement from human evaluation \label{appendix: iaa}}

To assess annotation consistency, we measured pairwise agreement among the four expert annotators (US1, US2, CA, and IN) for each evaluation metric. For every annotator pair, we computed Kendall, Pearson, and Spearman correlations, and report their average as the agreement score. Tables~\ref{tab:iaa-completeness}--\ref{tab:iaa-overall} show the raw pairwise agreement values. Table~\ref{tab:iaa-aggregated} summarizes the average inter-annotator agreement for each metric by averaging across all annotator pairs, along with an overall average across metrics.

Overall, agreement is moderate across metrics, with slightly higher consistency for completeness compared to factual accuracy, which remains the most challenging dimension for annotators.

\begin{table*}[htbp]
\centering
\caption{Raw pairwise inter-annotator agreement for the evaluation metrics. Each score is the average of Kendall, Pearson, and Spearman correlations for the corresponding annotator pair.}
\label{tab:iaa-raw}
\begin{subtable}[t]{0.48\textwidth}
\centering
\caption{Completeness}
\label{tab:iaa-completeness}
\begin{tabular}{lccr}
\toprule
Metric & Ann1 & Ann2 & Mean \\
\midrule
completeness & US1 & US2 & 0.593 \\
completeness & US1 & CA  & 0.510 \\
completeness & US1 & IN  & 0.659 \\
completeness & US2 & CA  & 0.545 \\
completeness & US2 & IN  & 0.575 \\
completeness & CA  & IN  & 0.483 \\
\bottomrule
\end{tabular}
\end{subtable}
\hfill
\begin{subtable}[t]{0.48\textwidth}
\centering
\caption{Factual accuracy}
\label{tab:iaa-factual}
\begin{tabular}{lccr}
\toprule
Metric & Ann1 & Ann2 & Mean \\
\midrule
factual accuracy & US1 & US2 & 0.576 \\
factual accuracy & US1 & CA  & 0.491 \\
factual accuracy & US1 & IN  & 0.456 \\
factual accuracy & US2 & CA  & 0.562 \\
factual accuracy & US2 & IN  & 0.531 \\
factual accuracy & CA  & IN  & 0.507 \\
\bottomrule
\end{tabular}
\end{subtable}

\vspace{0.8em}

\begin{subtable}[t]{0.48\textwidth}
\centering
\caption{Overall}
\label{tab:iaa-overall}
\begin{tabular}{lccr}
\toprule
Metric & Ann1 & Ann2 & Mean \\
\midrule
overall & US1 & US2 & 0.568 \\
overall & US1 & CA  & 0.514 \\
overall & US1 & IN  & 0.591 \\
overall & US2 & CA  & 0.575 \\
overall & US2 & IN  & 0.569 \\
overall & CA  & IN  & 0.477 \\
\bottomrule
\end{tabular}
\end{subtable}
\end{table*}

\begin{table}[htbp]
\centering
\caption{Aggregated inter-annotator agreement. For each metric, we report the average pairwise agreement across all annotator pairs. The final row reports the average across the remaining metrics.}
\label{tab:iaa-aggregated}
\begin{tabular}{lc}
\toprule
Metric & Avg.\ agreement \\
\midrule
Completeness     & 0.561 \\
Factual accuracy & 0.521 \\
Overall          & 0.549 \\
\midrule
Average across metrics & 0.544 \\
\bottomrule
\end{tabular}
\end{table}

\clearpage

\subsection{Failure Mode Analysis with Qualitative Examples \label{appendix: error}}
This section shows the qualitative examples of the error cases from omission of findings and false predictions from \ours, GPT-5, MedVersa and MedGemma.

\begin{figure}[!ht]
\centering
\small
\includegraphics[width=\textwidth]{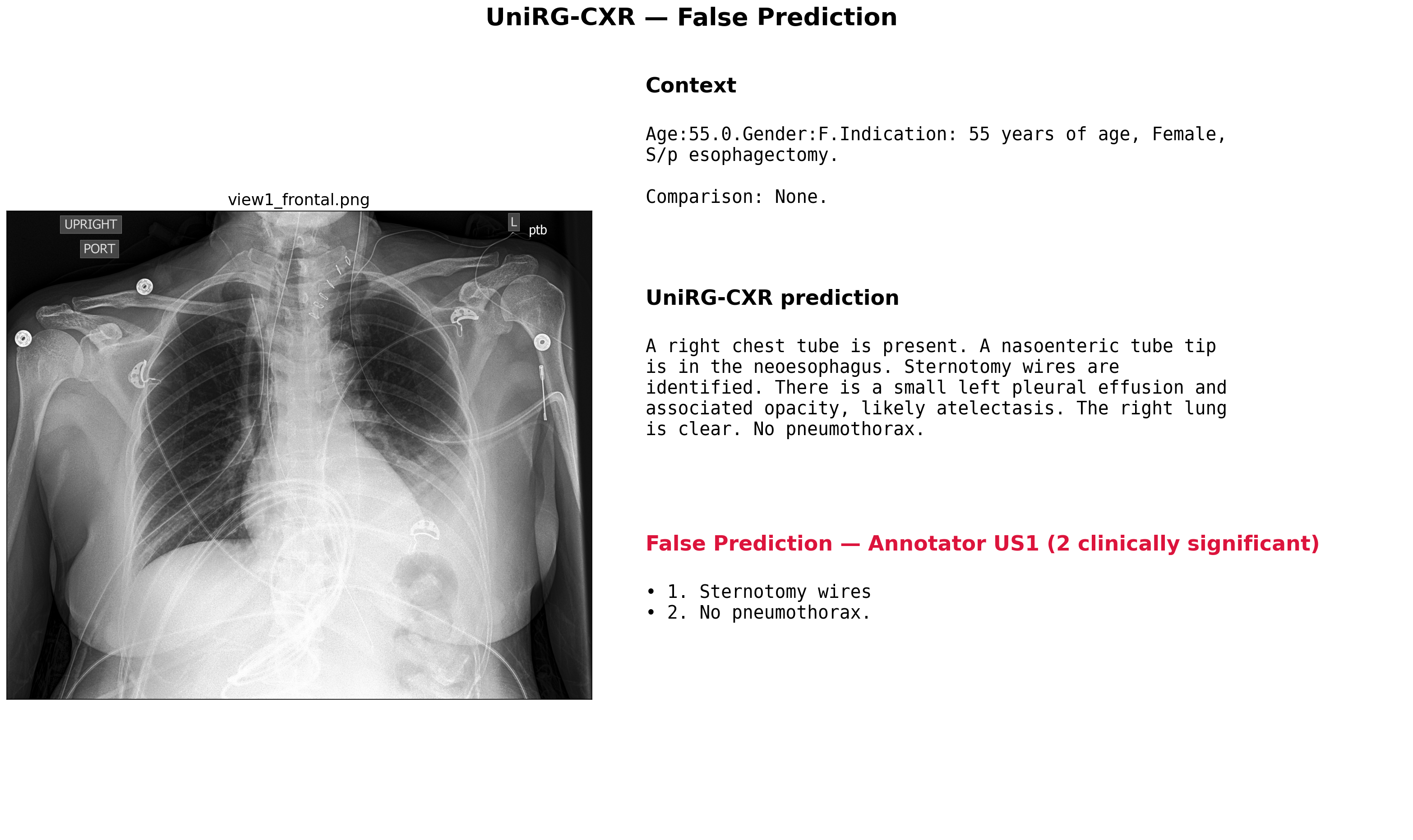}
\end{figure}

\begin{figure}[!ht]
\centering
\small
\includegraphics[width=\textwidth]{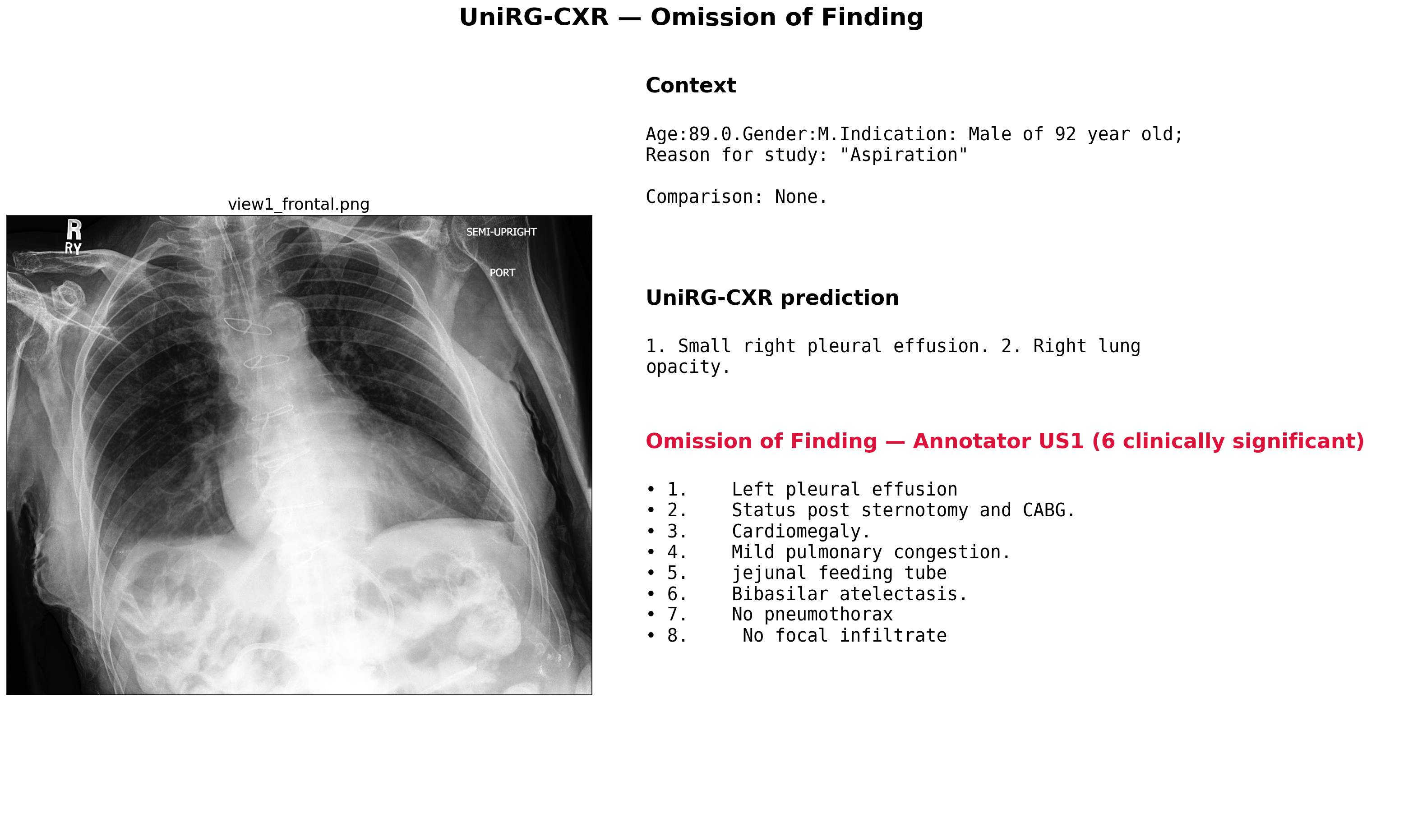}
\end{figure}

\begin{figure}[!ht]
\centering
\small
\includegraphics[width=\textwidth]{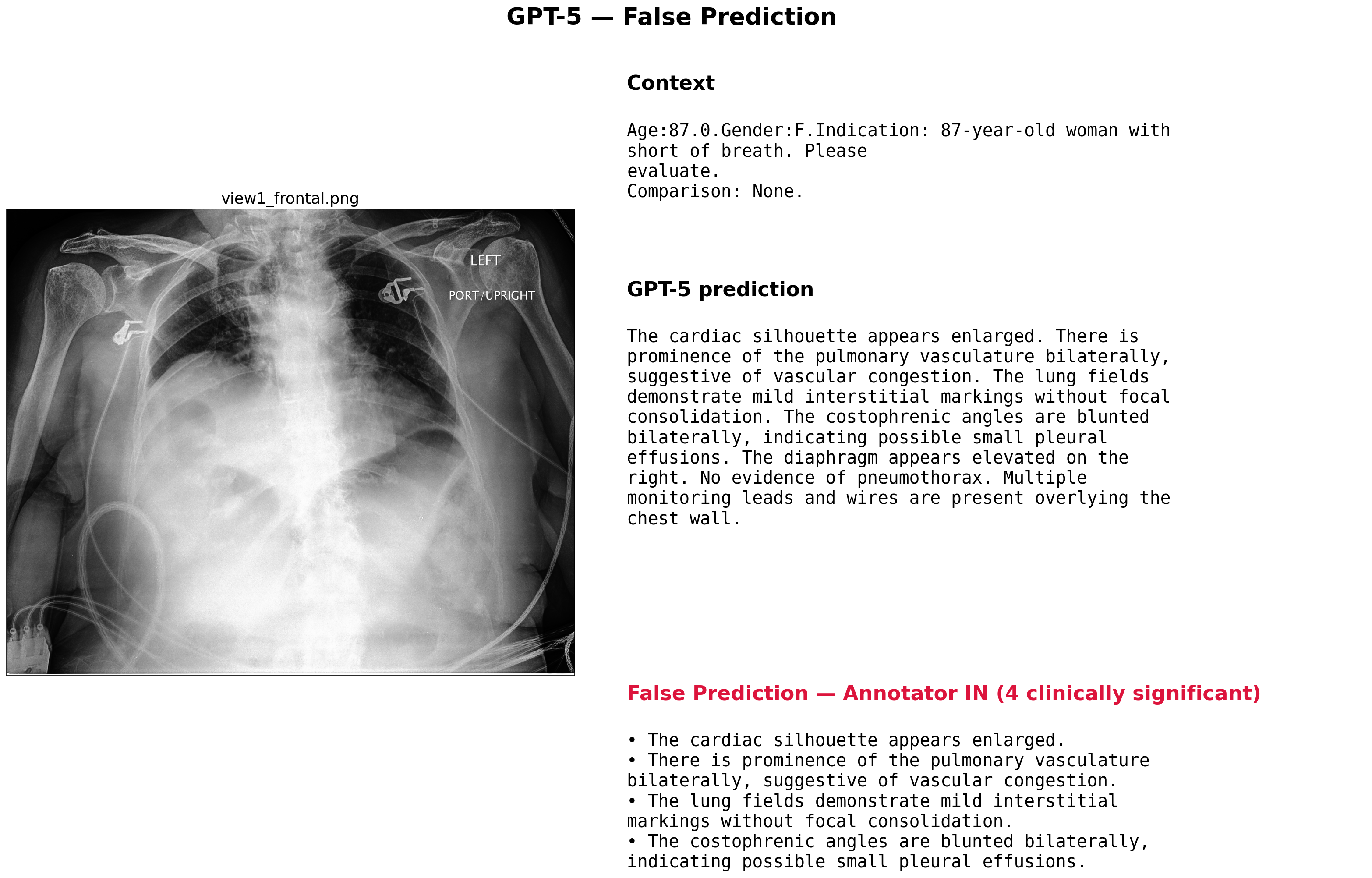}
\end{figure}

\begin{figure}[!ht]
\centering
\small
\includegraphics[width=\textwidth]{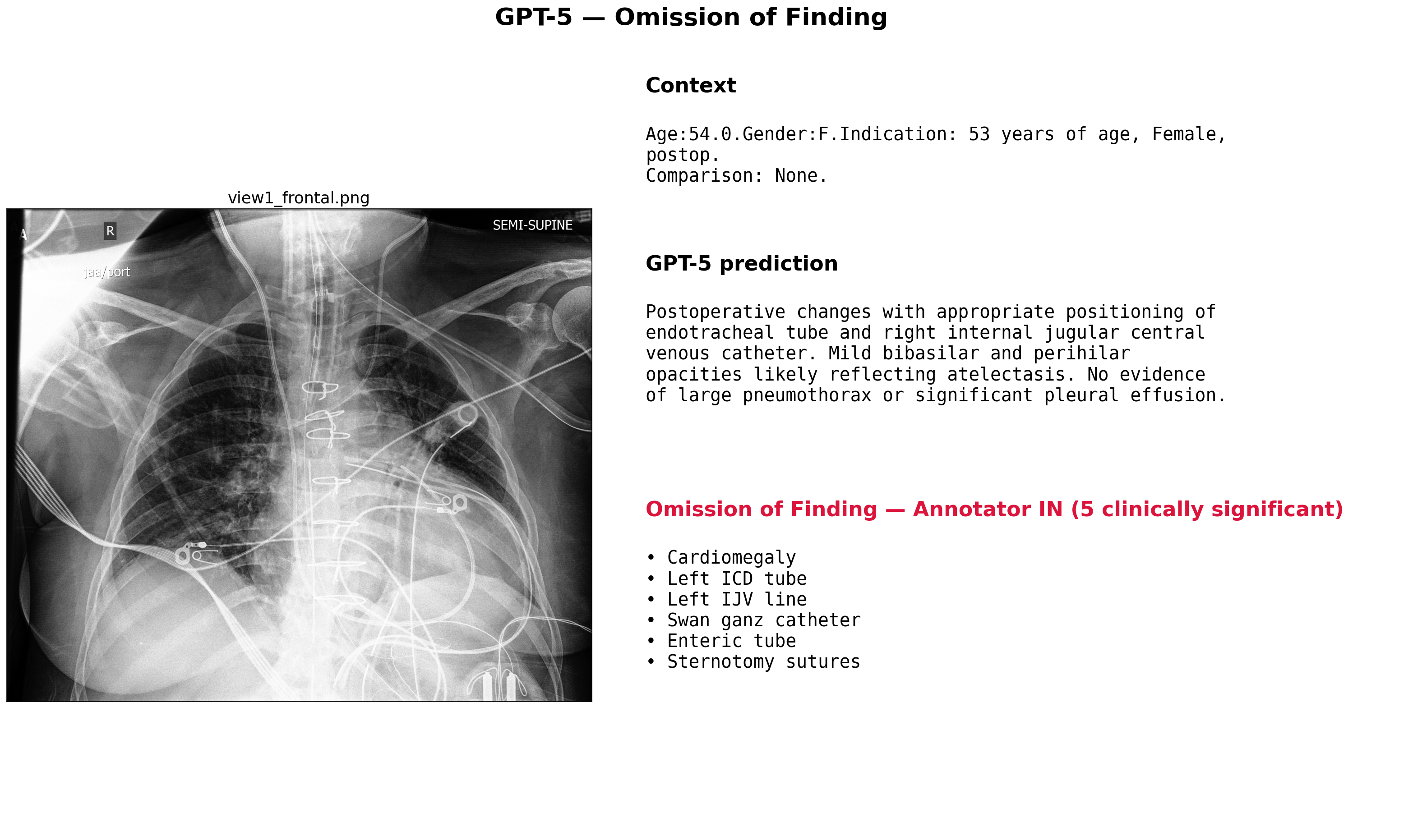}
\end{figure}

\begin{figure}[!ht]
\centering
\small
\includegraphics[width=\textwidth]{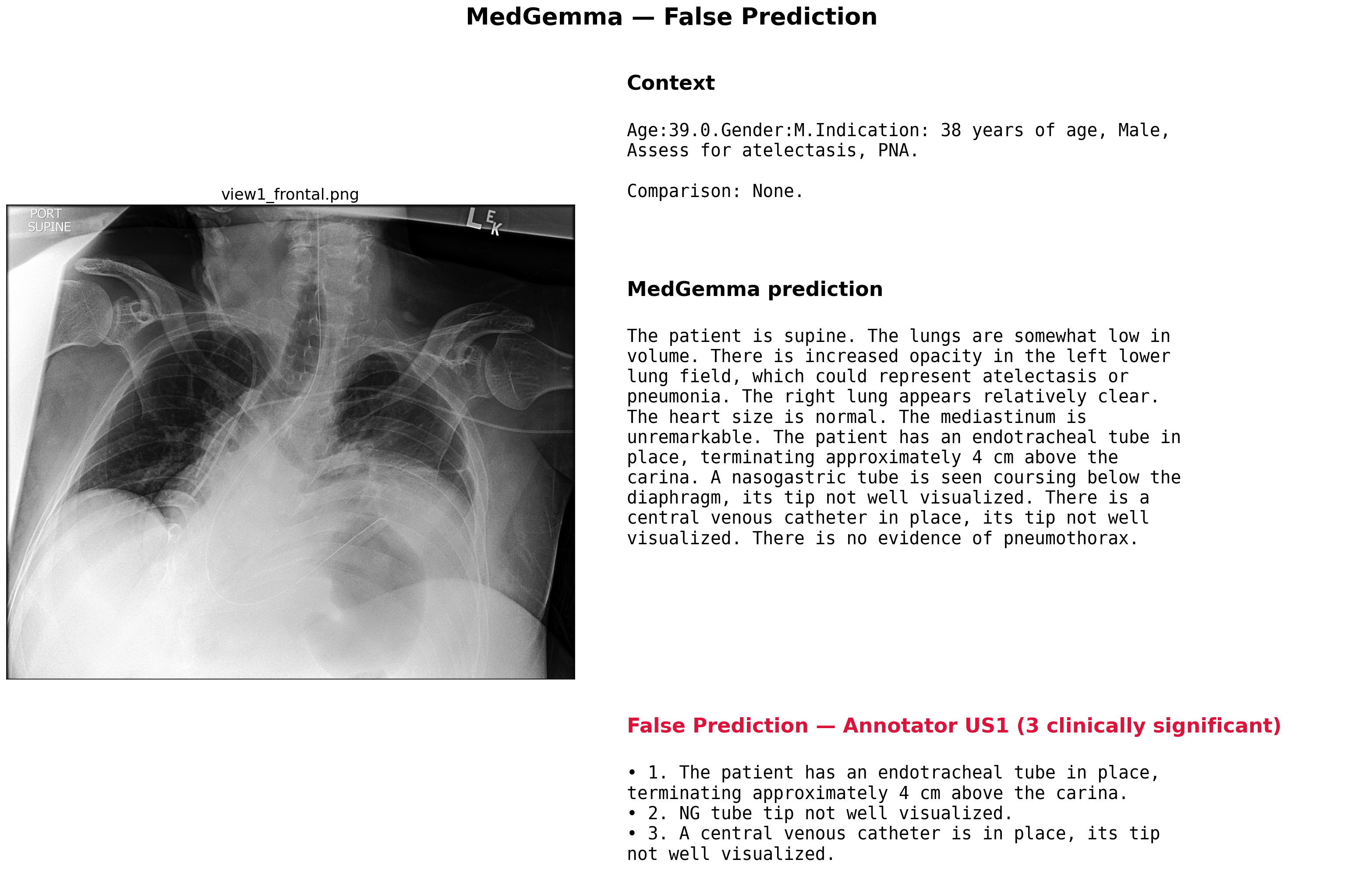}
\end{figure}

\begin{figure}[!ht]
\centering
\small
\includegraphics[width=\textwidth]{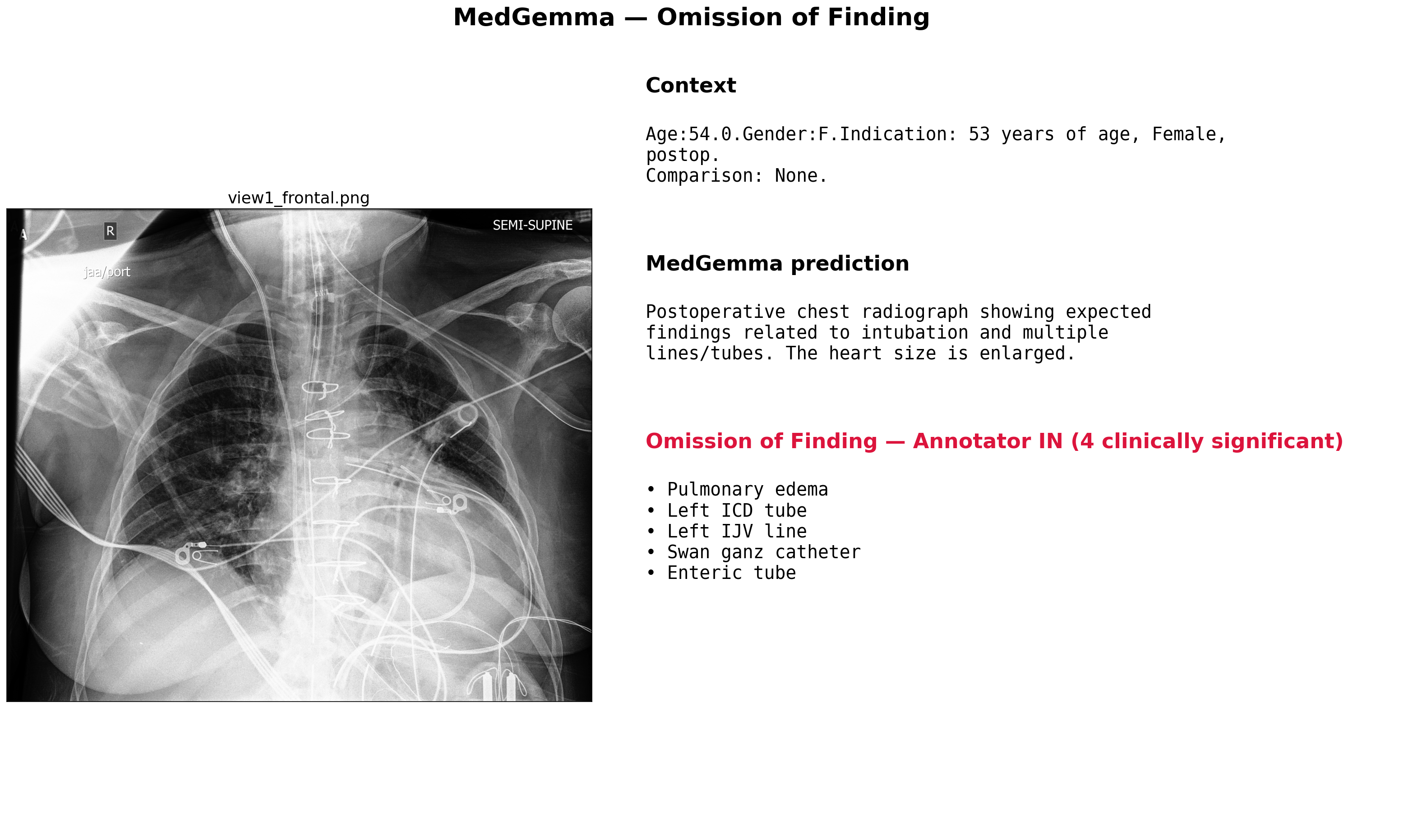}
\end{figure}

\begin{figure}[!ht]
\centering
\small
\includegraphics[width=\textwidth]{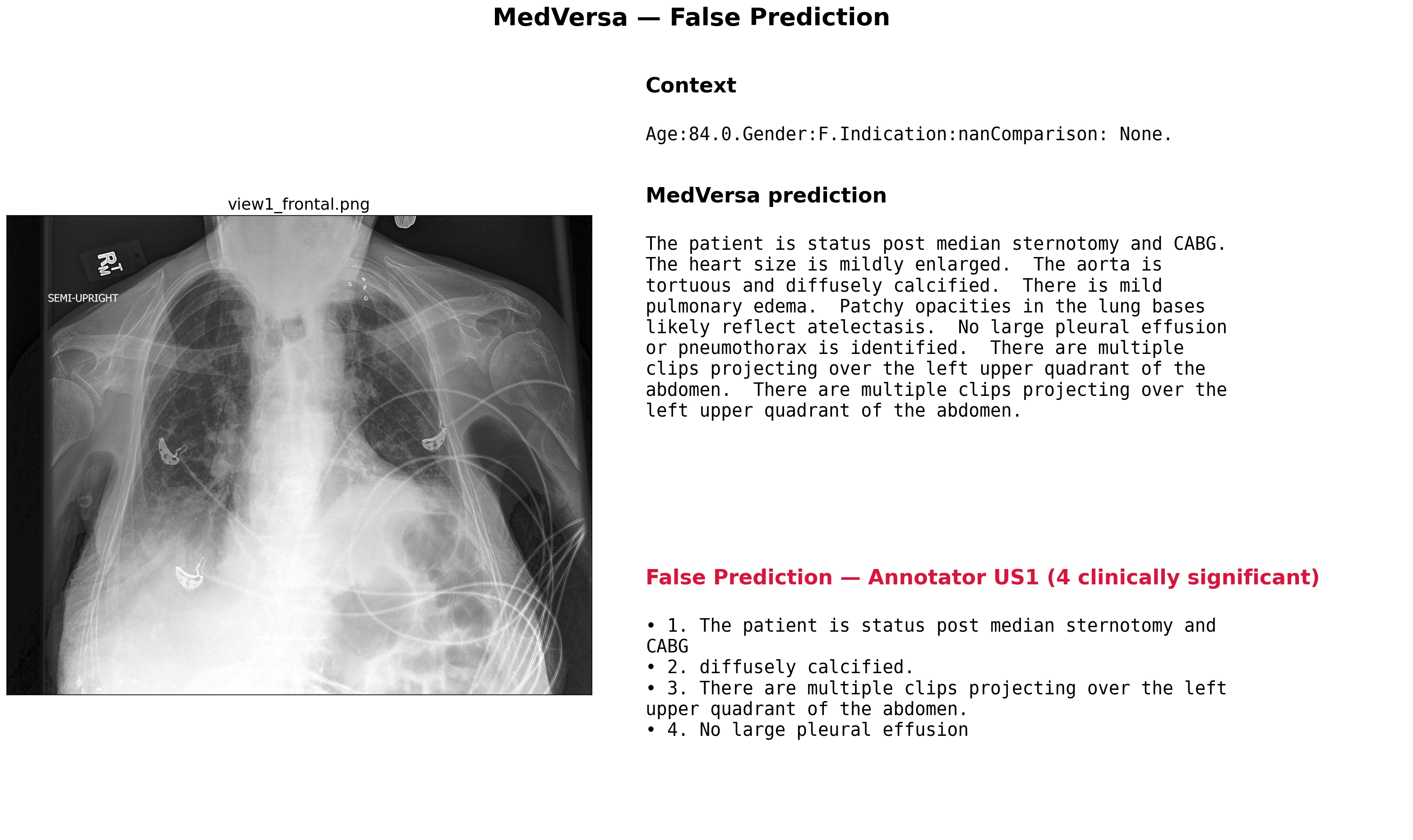}
\end{figure}

\begin{figure}[!ht]
\centering
\small
\includegraphics[width=\textwidth]{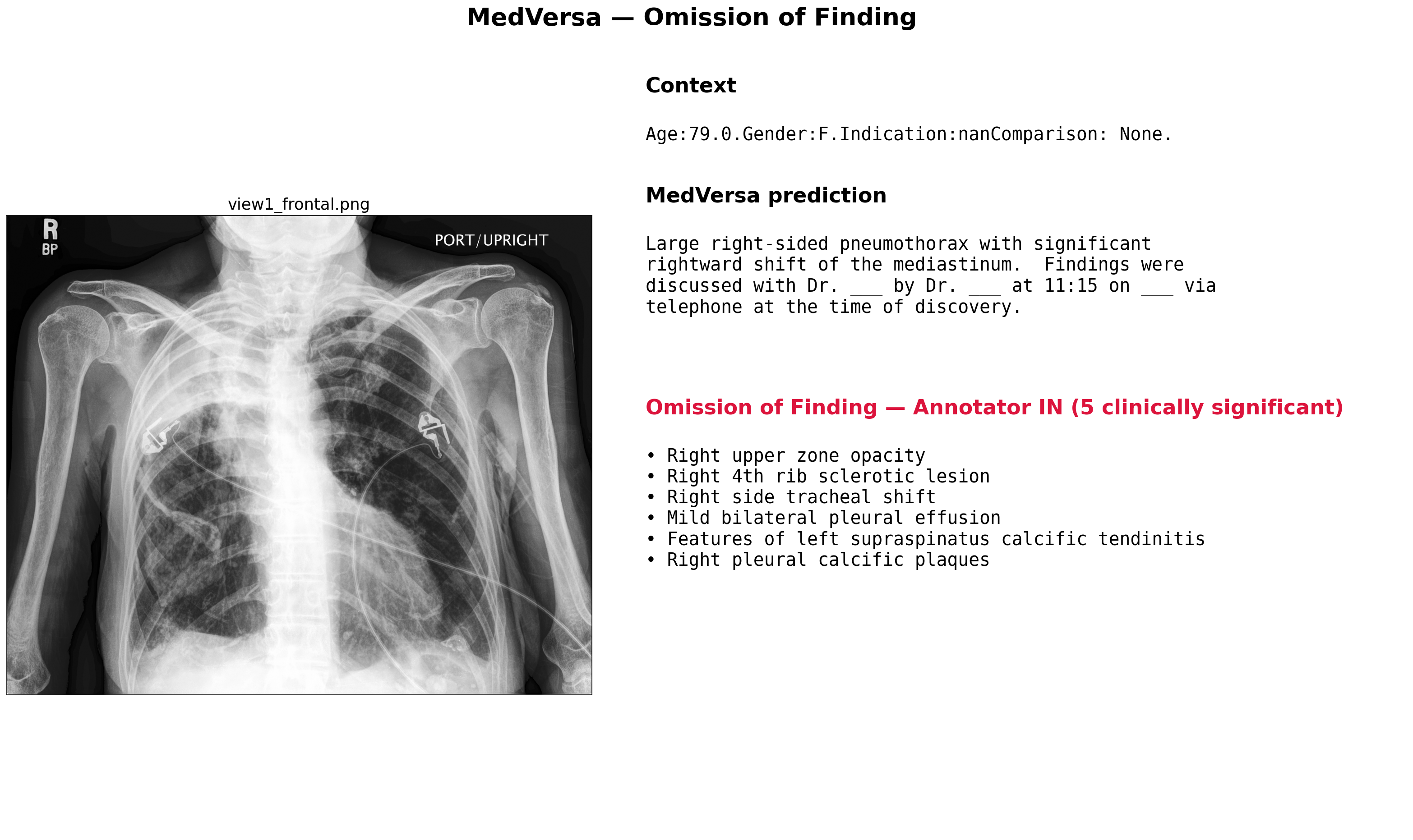}
\end{figure}

\end{document}